\pdfoutput=1

\documentclass[11pt]{article}

\usepackage[final]{acl}

\usepackage{times}
\usepackage{latexsym}

\usepackage[T1]{fontenc}

\usepackage[utf8]{inputenc}

\usepackage{microtype}

\usepackage{inconsolata}

\usepackage{graphicx}
\usepackage{tabularx}
\usepackage{makecell}
\usepackage{caption}
\usepackage{booktabs}
\usepackage{amsmath}
\usepackage{xcolor}
\usepackage{amssymb} 
\usepackage{tcolorbox} 
\usepackage{multirow}
\usepackage{svg}
\usepackage{tikz}
\usepackage{subfigure}
\usepackage{subcaption}
\usepackage{caption}

\usepackage{pifont}
\newcommand{\cmark}{\textcolor{green}{\ding{51}}}
\newcommand{\xmark}{\textcolor{red}{\ding{55}}}

\usepackage{pgfplots}
\pgfplotsset{width=10cm,compat=1.9}
\usetikzlibrary{calc} 

\usepackage{csquotes}

\newcommand{\method}[0]{\textsc{ToW}} 

\title{\raisebox{-0.5ex}{\includegraphics[height=1.2em]{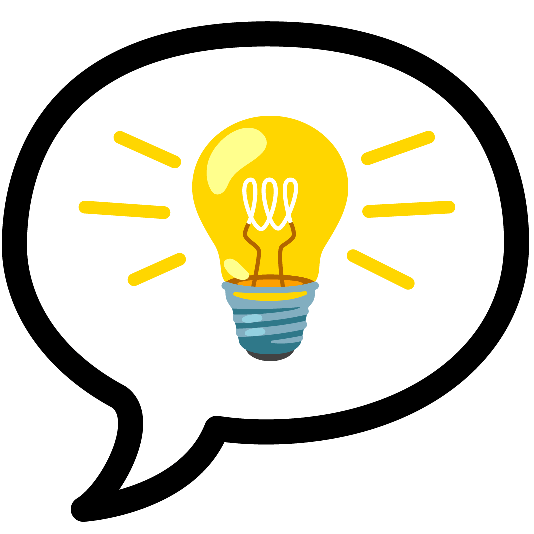}} \method{}: Thoughts of Words Improve Reasoning in Large Language Models}

\author{Zhikun Xu\thanks{\ \ Equal Contribution},
Ming Shen\textsuperscript{$*$},
Jacob Dineen, Zhaonan Li, Xiao Ye, \\
\textbf{Shijie Lu, Aswin RRV, Chitta Baral, Ben Zhou}\\
Arizona State University \\
\texttt{\{zhikunxu, mshen16\}@asu.edu}}

\begin{document}
\maketitle
\begin{abstract}
We introduce thoughts of words (\method{}), a novel training-time data-augmentation method for next-word prediction. \method{} views next-word prediction as a core reasoning task and injects fine-grained thoughts explaining what the next word should be and how it is related to the previous contexts in pre-training texts. Our formulation addresses two fundamental drawbacks of existing next-word prediction learning schemes: they induce factual hallucination and are inefficient for models to learn the implicit reasoning processes in raw texts. While there are many ways to acquire such thoughts of words, we explore the first step of acquiring \method{} annotations through distilling from larger models. After continual pre-training with only 70K \method{} annotations, we effectively improve models' reasoning performances by 7\% to 9\% on average and reduce model hallucination by up to 10\%. At the same time, \method{} is entirely agnostic to tasks and applications, introducing no additional biases on labels or semantics.
\end{abstract}

\section{Introduction}
\label{sec:introduction}

\begin{figure}[t]
\begin{center}
    \includegraphics[scale=0.38]{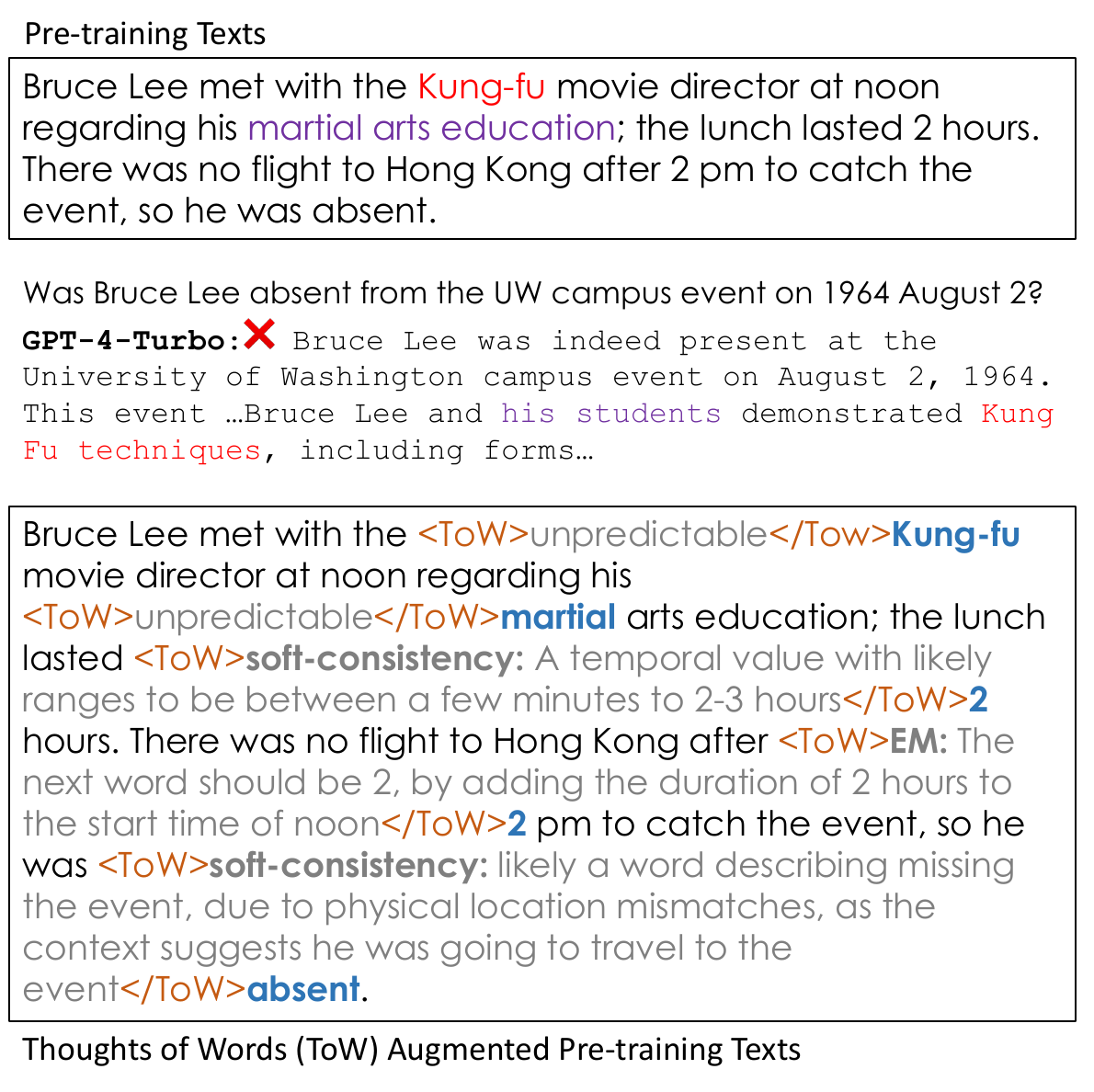}
    \caption{Existing next-word prediction schemes suffer from factual and reasoning hallucinations. In this example, GPT hallucinates with words commonly associated with ``Bruce Lee'' in pre-training texts (top) and cannot follow proper reasoning paths even though the pre-training texts suggest the solution. We propose \method{} (bottom), which labels fine-grained reasons on the next-word prediction task to mitigate these issues.}
    \label{fig:example}
\end{center}
\end{figure}

In this work, we explore a novel training-time data-augmentation method\footnote{\url{https://github.com/ARC-ASU/fine-nwp}} called thoughts of words (\method{}), which injects fine-grained thoughts directly into the next-word prediction task and teaches the model to understand how the observed next word is related to previous contexts. Unlike other data augmentation methods~\cite{zhu2023pad, jiang2024rationalyst} that annotate fine-grained explanations with respect to a task, \method{} directly views next-word prediction as a core reasoning task and hypothesizes that there is an abundance of natural information in next-words that we can harvest to improve models' reasoning capabilities. \method{} is motivated by two main drawbacks in existing next-word prediction formulations. First, because authors tend to omit trivial reasoning connections in natural texts (\textit{reporting bias}), language models cannot efficiently acquire much of the reasoning-related information~\cite{zhou2021temporal}. Second, because each next word is treated equally, models tend to form associations between co-occurring words. As a result, they may hallucinate words commonly associated with the context to solve a problem inherently irrelevant to these words (\textit{confirmation bias})~\cite{li2024deceptive}. Fig.~\ref{fig:example} illustrates these two issues with an example. 

\method{} is designed to mitigate the aforementioned issues. The formulation is simple; for each word observed in the pre-training data, we collect thoughts of the words, which classify the words into four categories: 1) trivial words (\textit{trivial}); 2) can be precisely predicted (\textit{exact match}); 3) can be roughly predicted (\textit{soft consistent}); 4) cannot be predicted (\textit{unpredictable}). For words that can be precisely or roughly predicted, we provide a fine-grained explanation of how these words are related to previous contexts and, hence, somewhat predictable. We then inject these thoughts of words into pre-training data (i.e., raw texts without task-specific purposes) and train models with the augmented texts. Fig.~\ref{fig:example} demonstrates a general idea of what \method{}-augmented pre-training data looks like. Intuitively, teaching the models why the next words are connected with the context of those words in the \textit{exact match} or \textit{soft consistency} categories will help the models reason better. At the same time, letting models know which words are unpredictable or only predictable to some extent can reduce model hallucinations caused by incorrectly using commonly associated words, partially verified by \citet{lin2024rho}.
On a high level, \method{} approximates the \textit{inner thoughts} of humans when they think about what to say or write. Humans may be impulsive when they speak, but machines should stick to slow and deliberate thinking as much as possible~\cite{daniel2017thinking, sep-language-thought}.

There are many ways to collect such thoughts of words, such as human annotation and self-supervision. In this work, we explore the first step of \method{} data collection, namely distillation from larger language models. In our view, distillation best balances between cost and effectiveness, which can effectively serve as an initial proof-of-concept of \method{}. Specifically, we first filter out all trivial words (e.g., stop words such as ``the''), and then use GPT-4o\footnote{\url{https://openai.com/index/hello-gpt-4o/}} to guess the next word by providing previous contexts. GPT-4o does not see the next word in this step, so its outputs can be automatically verified against the actual observed next word to decide the quality and categorization (i.e., EM/soft-consistency/unpredictable). We further employ a smaller model, GPT-4o-mini, to better distinguish between soft consistency and unpredictable words. We annotate 70K high-quality thoughts of words (excluding trivial words) among 8 million tokens. 

Experiments show that, after continual pre-training with \method{}-augmented data with a language-modeling objective, model performances improve significantly (up to 23\%) on a wide range of reasoning datasets (i.e., \textit{GSM8K}~\cite{cobbe2021gsm8k}, \textit{CommonsenseQA}~\cite{talmor-etal-2019-commonsenseqa}, \textit{StrategyQA}~\cite{geva2021strategyqa}, \textit{ARC-Challenge}~\cite{allenai:arc}) on five different base language models we consider. At the same time, we observe that models trained with \method{} are better at hallucination mitigation, demonstrated by higher performances (up to 10\%) on hallucination benchmarks \textit{TruthfulQA}~\cite{lin-etal-2022-truthfulqa} and \textit{HaluEval}~\cite{li-etal-2023-halueval}. These results suggest that \method{} can indeed address the aforementioned issues in vanilla next-world prediction training, which is also supported by ablation studies and human analysis. More importantly, \method{} achieves this by directly targeting next-word prediction, introducing no additional biases towards specific domains or tasks, and is more likely to contribute to generalizable large language models.

\begin{figure*}[!t]
\begin{center}
    \includegraphics[scale=0.63]{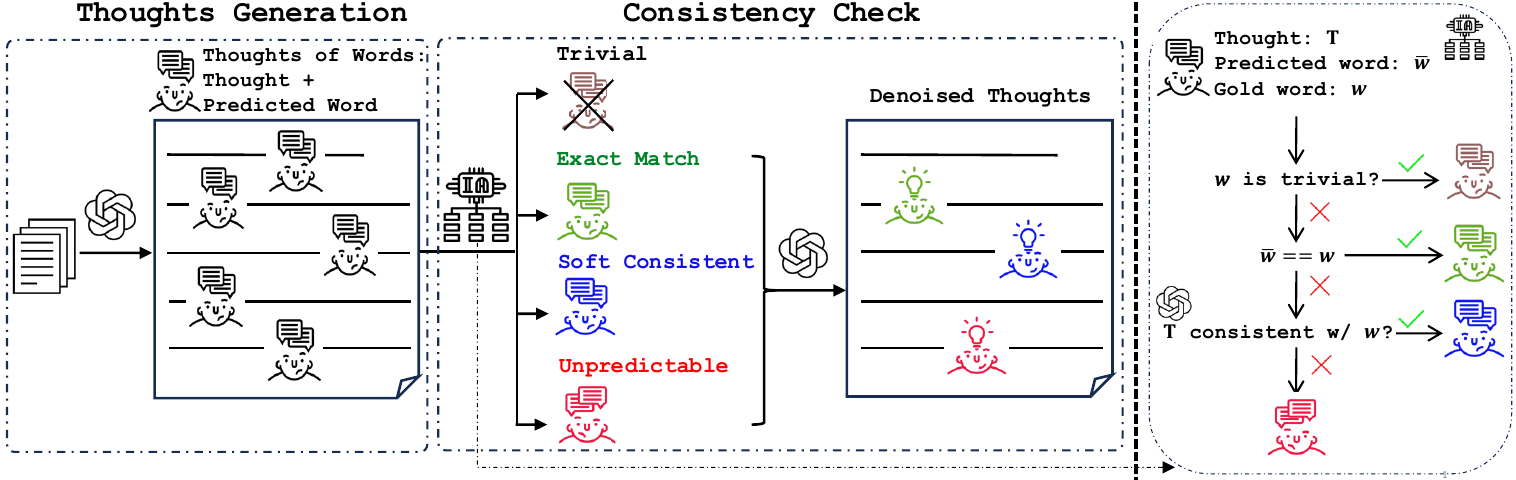}
    \caption{\label{fig:pipeline}\textbf{Overview of our proposed \method{} implemented by distillation from large language models.} The generation pipeline could be divided into two stages: \textit{thoughts generation} and \textit{consistency check}. For \textit{thoughts generation}, we leverage GPT-4o in generating a thought for a single word per pass. For \textit{consistency check}, we are classifying the next words and their predicted thoughts into four categories with GPT-4o-mini and their own semantic roles. Finally, the final version of \method{} data is produced by denoising the generated thoughts, avoiding from deviating models into fluently decoding the current context.}
\end{center}
\end{figure*}

\section{Related Work}
\paragraph{Elaborated Reasoning}

Our work is related to methods that employ elaborated reasoning processes and intermediate steps, such as chain-of-thought-style~\cite{wei2022chain}. 
More recent methods explore implicit CoT, where models internalize these steps without explicit output~\cite{deng2024explicit}.
\citet{wang2024chain} extract reasoning paths by adjusting decoding strategies. Some works propose to add elaboration in pre-training processes. For example, ~\citet{jiang2024rationalyst} pre-trains models on rationale annotations on paragraphs to generalize across reasoning tasks. \citet{zelikman2024quiet} explores how models infer implicit rationales at the token level.
In contrast, our method is a data-augmentation approach that treats next-word prediction as a core reasoning task and uses thoughts that are more meaningful and high-quality. Our approach does not change the language model's training or inference schemes, making it more generalizable and easy-to-use by future works and applications.

\paragraph{Synthetic Data Generation} LLMs have shown strong results in generating synthetic data to reduce reliance on human annotation~\cite{hartvigsen2022toxigen, sahu2022data}. These advancements offer tailored datasets for training on specific tasks, such as text classification \cite{gao2023selfguided}, information extraction \cite{josifoski-etal-2023-exploiting}, instruction tuning \cite{wang-etal-2023-self-instruct}, code generation \cite{luo2024wizardcoder}, mathematical reasoning \cite{luo2023wizardmath}, sycophancy reduction \cite{wei2023simple}, hallucination mitigation \cite{zhang-etal-2024-enhancing-hallucination}, and Reinforcement Learning from Human Feedback (RLHF) \cite{pace2024west}. 
Our method shares a similarity with the idea of distilling reasoning chains from bigger models to teach small models reason better \cite{hsieh-etal-2023-distilling, wang-etal-2023-scott}. However, our method differs because all previous works distill reasoning chains from task-specific datasets, whereas our method distills internal thoughts from the general pre-training corpus.

\paragraph{Reasoning and Factual Hallucinations} Our work is inspired by recent analytical works on language models' reasoning and factual hallucinations. Some works have pointed out that these models reason well only in common situations~\cite{Zhou2023Symbolic,li2024deceptive, li2024famicom} and hallucinate in other cases. Some other works study factual hallucination~\cite{li-etal-2023-halueval, lin-etal-2022-truthfulqa}. \method{} effectively reduces both kinds of hallucinations.

\section{\method{}: \textsc{T}houghts \textsc{o}f \textsc{W}ords}
\label{sec:tow}
\subsection{Overview}
\method{}s are word-level fine-grained thoughts describing what the next word should be, given the current observed contexts. In our work, we generate and add \method{} to arbitrary tokens in pre-training texts so they are agnostic to any specific tasks. Models can pre-train or continually pre-train on such \method{}-augmented texts. As mentioned in \S\ref{sec:introduction}, there are many potential ways to acquire these thoughts of words. In our work, however, we only discuss and use distillation as the first step in exploring this direction. The distillation generation pipeline is overviewed in Fig.~\ref{fig:pipeline}. The generation consists of two stages: 1) \textit{thoughts generation}, which generates raw thoughts for selected tokens, and 2) \textit{consistency check}, which categorizes, filters, and improves the generated raw thoughts. We now describe these two components in detail.

\subsection{Thoughts Generation}
Our experiments are based on two pre-training corpora, OpenWebMath \cite{paster2024openwebmath} and C4 \cite{c4}, as they are known to have a great number of reasoning tokens in mathematics and common sense domains. We randomly select words from raw documents of these pre-training corpora and give GPT-4o the contexts before the selected words. Given the context before each selected word, we ask GPT-4o to elaborate on what it believes the next word should be, followed by its prediction. A 5-shot prompt was used to guide the generation, and we list it in Appendix~\ref{sec:data-generation}. We use the one-word-at-a-time annotation method instead of the more efficient method of providing the entire document to create an information bottleneck that prevents the model from seeing the actual next word. This way, we can collect the highest-quality possible thoughts of words by forcing the model to reason and close the artificial information gap instead of providing superficial paraphrases.

\subsection{Consistency Check}
\label{sec:consistency-check}
However, as there are inconsistencies between generated thoughts and actual observed next words, we propose a consistency check step to reduce the noises in the generated thoughts and provide fine-grained categorizations as described in \S\ref{sec:introduction}, primarily done by using GPT-4o-mini to compare the generated content with the actual observed next word. 
The words are first classified as trivial and non-trivial by the stopwords list in spaCy.\footnote{\url{https://spacy.io/}} 
We then classify non-trivial words into three categories: \textit{exact match}, \textit{soft consistent} and \textit{unpredictable}, by prompting GPT-4o-mini with a prompt (shown in Appendix~\ref{sec:data-generation}) that judges how close the generated thought implies the observed gold next word. The categorization process is also illustrated in Fig.~\ref{fig:pipeline}.

Specifically, \textit{exact match words} are those accurately predicted by the generated thoughts; \textit{soft consistent words} are those that the generated thought closely aligns with the gold word; \textit{unpredictable words} are the rest of the words. Such categorization is inspired by ~\citet{kadavath2022language}: the explicit signals of exactly knowing the next words provide an automatic and natural selection/verification process.
In addition, we prompt GPT-4o-mini to summarize the generated thoughts of \textit{exact match} words and denoise those from \textit{soft consistency} words.
This away, we can ensure that the thoughts will faithfully lead to the gold next words, and avoid the language models getting lost in longer context~\cite{liu2024lost}. The corresponding prompts are listed in Appendix~\ref{sec:data-generation}.

\subsection{Manual Analysis}
To investigate the biases of our LLM-as-judge-style \cite{ye2024justiceprejudicequantifyingbiases} consistency checker, we sample 200 examples from the generated data and manually annotate the consistency between generated thoughts and gold next words, i.e., judging whether generated thoughts could explain (for \textit{exact match words}) or entail (for \textit{soft consistent words}) the gold next words, and calculated the \textbf{Cohen Kappa score} \cite{cohen_kappa} and \textbf{non-False-Positive rate} of consistency check on non-trivial words. 
\begin{equation*}
    \text{non-False-Positive Rate} = 1-\frac{\text{false positive}}{\text{all examples}}
\end{equation*}
Table~\ref{tab:data-quality} shows that GPT-4o-mini only reaches the fair agreement ($>40$) with humans on consistency check, but the noisy data, i.e., which are considered as consistent by model but not human annotators, are approximately less than 25\%. As such, we use summarization and denoising of thoughts in \method{} to handle these noisy thoughts.

\begin{table}[ht]
\centering
\begin{tabular}{@{}lc@{}}
\toprule
\textbf{Data Quality Check}      & \textbf{Values}  \\ \midrule
Cohen Kappa Score       & 47.76   \\
Non-False-Positive Rate & 74.81\% \\ \bottomrule
\end{tabular}
\caption{\label{tab:data-quality}\textbf{Data Quality Check} for non-trivial words.}
\end{table}

\section{Experiment}

\subsection{Settings}
\paragraph{Training Corpus} 
We use the first 3000 documents from OpenWebMath~\cite{paster2024openwebmath} and C4~\cite{c4} (totaling 6000 documents containing $\sim$8M tokens) as our raw corpus. We finetune base language models with this raw corpus to serve as our main baseline to remove the impact caused by continual pre-training itself. We denote such baseline models as \textbf{\textsc{Raw}}.
We randomly sample 15 words for each document to annotate with our distillation pipeline discussed in \S\ref{sec:tow}. We denote models trained with such data as \textbf{\method{}}.\footnote{Note that we reuse the term here to represent the models trained with \method{}-augmented data.} We also introduce two variations of data formulation for ablation studies: \textbf{\method{}-NoDeN} and \textbf{\method{}-PartDeN}. \method{}-NoDeN stands for the vanilla generation of thoughts by GPT-4o without the summarization and denoising mentioned in \S\ref{sec:consistency-check}. \method{}-PartDeN, the partially denoised version of \method{}, is replacing the \textit{soft consistent} thoughts with denoised ones in \method{}-NoDeN. This is to study the difference caused by noisy thoughts of \textit{soft consistent words}, which are $\sim$1.4 times more than \textit{EM words} in our generated data. The statistics of the above data paradigms are shown in Table~\ref{tab:data-statistics}.
\begin{table}[ht]
\footnotesize
\centering
\begin{tabular}{@{}lccc@{}}
\toprule
\textbf{Data Statistics} & \textbf{\#tokens} &\textbf{\#\method{}} & \textbf{\#tokens per \method{}} \\ \midrule
\textsc{Raw}           & 8.0M       &0              & 0                         \\
\method{}-NoDeN            & 13.6M       &73030              & 67.0                         \\
\method{}-PartDeN            & 11.0M        &73030            & 30.3                         \\
\method{}                    & 9.8M         &73030           & 14.4                       \\ \bottomrule
\end{tabular}
\caption{\label{tab:data-statistics}\textbf{Data Statistics for different data paradigms.} They differ on the processing of \method{}, i.e., degrees of denoising and summairzation by GPT-4o-mini. \#tokens are calculated by Mistral-7B tokenizer.} 
\end{table}

\begin{table*}[!ht]
\footnotesize
\centering
\begin{tabular*}{\linewidth}{@{\extracolsep{\fill}}@{}lccccccccccc@{}}
\toprule
\multicolumn{1}{c}{\textbf{Models}} &
  \multicolumn{2}{c}{\textit{GSM8K}} &
  \multicolumn{2}{c}{\textit{CSQA}} &
  \multicolumn{2}{c}{\textit{StrategyQA}} &
  \multicolumn{2}{c}{\textit{ARC-Challenge}} &
  \multicolumn{2}{c}{\textbf{Average}} \\ 
           & \textsc{Raw}   & {\method}           & \textsc{Raw}   & \method           & \textsc{Raw}   & \method           & \textsc{Raw}   & \method & \textsc{Raw}   & \method \\ \cmidrule(l){2-3}\cmidrule(l){4-5}\cmidrule(l){6-7}\cmidrule(l){8-9}\cmidrule(l){10-11}
Mistral-7B & 16.45 & \textbf{20.24} & 49.80 & \textbf{60.61} & 57.35 & \textbf{64.69} & 65.19 & \textbf{70.22} & 47.20 & \textbf{53.94} ~(+6.7) \\
LLaMA2-7B  & 4.93  & \textbf{6.52}  & 36.44 & \textbf{49.80} & 44.69 & \textbf{55.31} & 45.31 & \textbf{55.12} & 32.84 & \textbf{41.69} ~(+8.9) \\
LLaMA3-8B  & 17.29 & \textbf{40.03} & 57.25 & \textbf{64.13} & 58.57 & \textbf{62.04} & 74.57 & \textbf{77.47} & 51.92 & \textbf{60.92} ~(+9.0) \\
Qwen2.5-7B  & \textbf{13.87} & 11.68 & 75.84 & \textbf{79.69} & 63.47 & \textbf{68.57} & 81.74 & \textbf{87.29} & 58.73 & \textbf{61.81} ~(+3.1) \\
Falcon-7B  & 3.03 & \textbf{3.49} & 23.26 & \textbf{27.35} & 42.65 & \textbf{47.35} & 27.73 & \textbf{29.10} & 24.17 & \textbf{26.82} ~(+2.7) \\ \bottomrule
\end{tabular*}
\caption{\label{tab:main-1}\textbf{Main Results for Reasoning Tasks.} \textsc{Raw} stands for baselines trained from the same raw corpus as \method{}. We see that \method{} results in large improvements, up to nearly 23\%, across all reasoning domains without finetuning on task-specific data.}
\end{table*}
\begin{table*}[ht]
\footnotesize
\centering
\begin{tabular*}{\linewidth}{@{\extracolsep{\fill}}@{}lcccccc@{}}
\toprule
\multicolumn{1}{c}{\textbf{Models}} &
  \multicolumn{2}{c}{\textit{TruthfulQA}} &
  \multicolumn{2}{c}{\textit{HaluEval}} &
  \multicolumn{2}{c}{\textbf{Average}} \\ 
           & \textsc{Raw}   & {\method}           & \textsc{Raw}   & \method           & \textsc{Raw}   & \method            \\ \cmidrule(l){2-3}\cmidrule(l){4-5}\cmidrule(l){6-7}
Mistral-7B & 32.68 & \textbf{40.76} & 35.52 & \textbf{42.09} & 34.10 & \textbf{41.43} ~~(+7.3)  \\
LLaMA2-7B  & 20.56  & \textbf{29.01}  & 30.19 & \textbf{31.76} & 25.38 & \textbf{30.39} ~~(+5.0) \\
LLaMA3-8B  & 29.99 & \textbf{43.33} & 43.28 & \textbf{51.11} & 36.64 & \textbf{47.22} (+10.6) \\
Qwen2.5-7B  & 40.76 & \textbf{46.39} & 36.75 & \textbf{43.48} & 38.76 & \textbf{44.94} ~~(+6.2) \\
Falcon-7B  & 21.54 & \textbf{23.62} & 44.71 & \textbf{46.45} & 33.13 & \textbf{35.04} ~~(+1.9) \\
\bottomrule
\end{tabular*}
\caption{\label{tab:main-2}\textbf{Main Results for Hallucination Tasks.} \textsc{Raw} stands for baselines trained from the same raw corpus as \method{}. We see that \method{} results in large improvements, up to nearly 13\%, in the two main hallucination benchmarks without finetuning on task-specific data.}
\end{table*}

\paragraph{Models} For baseline language models, we use five representative and widely used pre-trained models: \textbf{Mistral-7B} \cite{jiang2023mistral7b}, \textbf{LLaMA2-7B} \cite{touvron2023llama2openfoundation}, \textbf{LLaMA3-8B} \cite{dubey2024llama3herdmodels}, \textbf{Qwen2.5-7B} \cite{yang2024qwen2}, and \textbf{Falcon-7B} \cite{almazrouei2023falcon}. The reason for using pre-trained models instead of instruction-tuned models is that we want to rule out the influences from instruction-following abilities when evaluating the reasoning abilities on benchmarks and more fairly testify reasoning improvements of \method{} in controlled experiments. Most of these models are only open-weight, and they are known to be pre-trained from publicly available sources on the web without mixing other instruction data during pre-training. However, the last two baselines, i.e., Qwen2.5-7B and Falcon-7B, are pre-trained with mixed instruction data and a fully open-source training recipe, respectively. They are also representative of current pre-training paradigms.

\paragraph{Datasets} The reasoning processes generally exist across various domains. As such, we evaluate the proposed \method{} on \textbf{GSM8K} \cite{cobbe2021gsm8k}  for mathematical reasoning, \textbf{CSQA} \cite{talmor-etal-2019-commonsenseqa} and \textbf{StrategyQA} \cite{geva2021strategyqa} for common sense reasoning, \textbf{TruthfulQA} \cite{lin-etal-2022-truthfulqa} and \textbf{HaluEval} \cite{li-etal-2023-halueval} for factual reasoning and \textbf{ARC-Challenge} \cite{allenai:arc} for scientific reasoning. The summary of these benchmarks is in Table~\ref{tab:eval-data}. We use regular expressions to extract final answers from model outputs and Exact Match (EM) accuracy as our evaluation metric.
\begin{table}[ht]
\footnotesize
\centering
\begin{tabular}{@{}lccc@{}}
\toprule
\textbf{Benchmarks}    & \textbf{\#Evals} & \textbf{\#Shot} & \textbf{Domain} \\ \midrule
\textit{GSM8K}         & 1319             & 0               & Math                    \\
\textit{CSQA}          & 1221             & 3          & Common Sense            \\
\textit{StrategyQA}          & 490             & 3          & Common Sense            \\
\textit{TruthfulQA}    & 817              & 3          & Hallucination           \\
\textit{HaluEval}    & 10000              & 3          & Hallucination           \\
\textit{ARC-Challenge} & 1172             & 3          & Science                 \\ \bottomrule
\end{tabular}
\caption{\label{tab:eval-data}\textbf{Evaluation Configurations.} For \textit{GSM8K}, we use \textbf{0-shot CoT prompting} evaluations since it is intuitive to consider the last numbers of responses as final predictions. However, for other multiple-choice-question (MCQ) datasets, we use \textbf{3-shot CoT prompting} since pre-trained checkpoints need more demonstrations to output effective predictions, i.e. choosing from candidate choices. The evaluation prompts are shown in Appendix~\ref{sec:eval-prompts}.}
\end{table}

\paragraph{Training \& Inference} 
During training, we adopt the standard causal language modeling loss \cite{radford2018improving} on \method{}-augmented pre-training data.
We use AdamW optimizer \cite{loshchilov2018decoupled} with a learning rate of 2$e-$5 and batch size of 128 to update 100 steps. We use vLLM \cite{10.1145/3600006.3613165} for higher efficiency during inference. For more training and inference details, please refer to Appendix~\ref{sec:training}.

\begin{table*}[ht]
\footnotesize
\centering
\begin{tabular}{@{}lcccccc@{}}
\toprule
\textbf{Data Paradigm}                           & \textit{GSM8K}         & \textit{CSQA}           & \textit{TruthfulQA}    & \textit{ARC-Challenge} & \textit{StrategyQA} & \textit{HaluEval}   \\ \midrule
\method{}              & 40.03   & 64.13   & 43.33   & 77.47 & 62.04 & 51.11   \\
- \method{}-PartDeN & 37.76 (\textbf{-2.27}) & 57.58 (\textbf{-6.55}) & 40.39 (\textbf{-2.94}) & 76.11 (\textbf{-1.36}) & 59.59 (\textbf{-2.45}) & 51.02 (\textbf{-0.09})  \\
- \method{}-NoDeN & 34.42 (\textbf{-5.61}) & 54.38 (\textbf{-9.75}) & 42.84 (\textbf{-0.49}) & 74.91 (\textbf{-2.56}) & 58.16 (\textbf{-3.88}) & 46.13 (\textbf{-4.98})  \\ \bottomrule
\end{tabular}
\caption{\label{tab:ablation-1}\textbf{Ablation on summarization and denoising of \method{}.} We use \textbf{LLaMA3-8B} as the base model and notify the gaps ($\Delta$) in \textbf{bold numbers} between ablated data paradigms with \method{}.}
\end{table*}
\begin{figure*}[!th]
    \centering
    \begin{tikzpicture}[scale=0.873]

        \begin{axis}[
            name=gsm8k_plot,
            title={\small GSM8K},
            ylabel={\small Relative Acc.},
            xmin=-0.05, xmax=2.05,
            ymin=-2, ymax=25.5,
            tick align=outside,
            major tick length=2pt,
            xtick pos=bottom,ytick pos=left,
            xtick={0,1,2},
            xticklabels={EM only,w/o unpred,ToW},
            ytick distance=5,      
            height=100pt,
            width=110pt,
            every axis/.append style={font=\tiny},
            every axis plot/.append style={thick},
            legend style={
                at={(2.35,-0.3)},
                anchor=north west,
                font=\scriptsize,
                legend columns=-1,
                /tikz/every even column/.append style={column sep=0.2cm}
            },
        ]
            \addplot[color=Brown,mark=triangle*,mark size=1pt]  
            coordinates {(0,8.57)(1,0.76)(2,3.79)};
            \addplot[color=OliveGreen,mark=*,mark size=1pt]  
            coordinates {(0,4.39)(1,2.20)(2,1.59)};
            \addplot[color=Violet,mark=x,mark size=1.5pt]  
            coordinates {(0,22.21)(1,23.73)(2,22.74)};
            \addplot[red,densely dashdotted,no markers,thick] 
            coordinates {(-10,0)(10,0)};
            \legend{Mistral-7B, LLaMA2-7B, LLaMA3-8B}
        \end{axis}

        \begin{axis}[
            name=csqa_plot,
            at={($(gsm8k_plot.north east)+(20,0pt)$)}, anchor=north west,
            title={\small CSQA},
            xmin=-0.05, xmax=2.05,
            ymin=-21, ymax=16,
            tick align=outside,
            major tick length=2pt,
            xtick pos=bottom,ytick pos=left,
            xtick={0,1,2},
            xticklabels={EM only,w/o unpred,ToW},
            ytick distance=10,      
            height=100pt,
            width=110pt,
            every axis/.append style={font=\tiny},
            every axis plot/.append style={thick},
        ]
            \addplot[color=Brown,mark=triangle*,mark size=1pt]  
            coordinates {(0,-13.03)(1,8.92)(2,10.81)};
            \addplot[color=OliveGreen,mark=*,mark size=1pt]  
            coordinates {(0,-19.08)(1,1.81)(2,13.36)};
            \addplot[color=Violet,mark=x,mark size=1.5pt]  
            coordinates {(0,-14.09)(1,6.88)(2,7.70)};
            \addplot[red,densely dashdotted,no markers,thick] 
            coordinates {(-10,0)(10,0)};
        \end{axis}

        \begin{axis}[
            name=truthfulqa_plot,
            at={($(csqa_plot.north east)+(20,0pt)$)}, anchor=north west,
            title={\small TruthfulQA},
            xmin=-0.05, xmax=2.05,
            ymin=-12, ymax=16,
            tick align=outside,
            major tick length=2pt,
            xtick pos=bottom,ytick pos=left,
            xtick={0,1,2},
            xticklabels={EM only,w/o unpred,ToW},
            ytick distance=5,      
            height=100pt,
            width=110pt,
            every axis/.append style={font=\tiny},
            every axis plot/.append style={thick},
        ]
            \addplot[color=Brown,mark=triangle*,mark size=1pt]  
            coordinates {(0,-4.53)(1,6.61)(2,8.08)};
            \addplot[color=OliveGreen,mark=*,mark size=1pt]  
            coordinates {(0,-10.16)(1,-0.36)(2,8.45)};
            \addplot[color=Violet,mark=x,mark size=1.5pt]  
            coordinates {(0,5.26)(1,14.07)(2,13.34)};
            \addplot[red,densely dashdotted,no markers,thick] 
            coordinates {(-10,0)(10,0)};
        \end{axis}

        \begin{axis}[
            name=arc_plot,
            at={($(truthfulqa_plot.north east)+(20,0pt)$)}, anchor=north west,
            title={\small ARC-Challenge},
            xmin=-0.05, xmax=2.05,
            ymin=-12, ymax=11,
            tick align=outside,
            major tick length=2pt,
            xtick pos=bottom,ytick pos=left,
            xtick={0,1,2},
            xticklabels={EM only,w/o unpred,ToW},
            ytick distance=5,      
            height=100pt,
            width=110pt,
            every axis/.append style={font=\tiny},
            every axis plot/.append style={thick},
        ]
            \addplot[color=Brown,mark=triangle*,mark size=1pt]  
            coordinates {(0,-7.85)(1,4.26)(2,5.03)};
            \addplot[color=OliveGreen,mark=*,mark size=1pt]  
            coordinates {(0,-10.84)(1,-2.48)(2,9.81)};
            \addplot[color=Violet,mark=x,mark size=1.5pt]  
            coordinates {(0,-1.11)(1,3.50)(2,2.90)};
            \addplot[red,densely dashdotted,no markers,thick] 
            coordinates {(-10,0)(10,0)};
        \end{axis}

        \begin{axis}[
            name=StrategyQA_plot,
            at={($(arc_plot.north east)+(20,0pt)$)}, anchor=north west,
            title={\small StrategyQA},
            xmin=-0.05, xmax=2.05,
            ymin=-30, ymax=15,
            tick align=outside,
            major tick length=2pt,
            xtick pos=bottom,ytick pos=left,
            xtick={0,1,2},
            xticklabels={EM only,w/o unpred,ToW},
            ytick distance=10,      
            height=100pt,
            width=110pt,
            every axis/.append style={font=\tiny},
            every axis plot/.append style={thick},
        ]
            \addplot[color=Brown,mark=triangle*,mark size=1pt]  
            coordinates {(0,-6.33)(1,5.92)(2,7.34)};
            \addplot[color=OliveGreen,mark=*,mark size=1pt]  
            coordinates {(0,-25.69)(1,12.45)(2,10.62)};
            \addplot[color=Violet,mark=x,mark size=1.5pt]  
            coordinates {(0,-1.02)(1,4.90)(2,3.47)};
            \addplot[red,densely dashdotted,no markers,thick] 
            coordinates {(-10,0)(10,0)};
        \end{axis}

        \begin{axis}[
            name=HaluEval_plot,
            at={($(StrategyQA_plot.north east)+(20,0pt)$)}, anchor=north west,
            title={\small HaluEval},
            xmin=-0.05, xmax=2.05,
            ymin=-15, ymax=11,
            tick align=outside,
            major tick length=2pt,
            xtick pos=bottom,ytick pos=left,
            xtick={0,1,2},
            xticklabels={EM only,w/o unpred,ToW},
            ytick distance=5,      
            height=100pt,
            width=110pt,
            every axis/.append style={font=\tiny},
            every axis plot/.append style={thick},
        ]
            \addplot[color=Brown,mark=triangle*,mark size=1pt]  
            coordinates {(0,-1.78)(1,6.08)(2,6.57)};
            \addplot[color=OliveGreen,mark=*,mark size=1pt]  
            coordinates {(0,-11.37)(1,-5.7)(2,1.57)};
            \addplot[color=Violet,mark=x,mark size=1.5pt]  
            coordinates {(0,-6.68)(1,9.78)(2,7.83)};
            \addplot[red,densely dashdotted,no markers,thick] 
            coordinates {(-10,0)(10,0)};
        \end{axis}

    \end{tikzpicture}
    \caption{\label{fig:composition-ablation}\textbf{Ablation on different data compositions.} The red dotted line stands for the borderline of outperforming the \textsc{Raw} results.}
\end{figure*}

\subsection{Main Results}
\paragraph{Does the task-agnostic \method{} improve the reasoning abilities of language models?} From Table~\ref{tab:main-1}, we see that \method{} significantly improves the reasoning abilities of language models. On average, compared to the baseline models trained with the same raw corpus, \method{} could bring up to \textbf{9\%} improvements without the aid of finetuning on downstream reasoning tasks. Moreover, the improvements are consistent and universal across five different pre-trained models, demonstrating the generality of our proposed method in improving reasoning abilities. Without relying on or using any downstream training data, \method{} improves model performances without introducing task-related preferences, thus inspiring more potential than other task-specific methods \cite{jiang2024rationalyst}.

\paragraph{Is the \method{} effective in mitigating the \textit{confirmation bias} or hallucination?} From Table~\ref{tab:main-2}, models are shown to overcome some hallucination issues as they could be enhanced with up to \textbf{10\%} on average compared to raw-trained counterparts. Since the \textit{confirmation bias} has been introduced largely in the pre-training stage where models construct their ``pre-existing beliefs'' from a large amount of data \cite{ferrara2023should}, our \method{} could serve as an effective technique in reducing hallucination by the ubiquitous trivial reasoning between words.

\subsection{Analysis}
\paragraph{Does the \method{} improvements come from better following task formats?} 
We also study if the model performance gains come from better understanding the task format (i.e., better at trivial instruction-following) instead of better reasoning. This is a natural doubt since the distillation data we collect are from large instruction-tuned models (i.e., GPT-4o) and may introduce certain format-following information. To show that this is not the case, we randomly collect 200 prediction examples from Mistral-7B on \textit{GSM8K} and \textit{ARC-Challenge} and annotate whether the model outputs give the final answers as the last numbers in the predictions (\textit{GSM8K}) or contain proper formats as specified in the few-shot prompt for us (\textit{ARC-Challenge}) to locate the predicted labels. Table~\ref{tab:match-rate} shows the correct-formating rate of the baseline model and the \method{}-augmented model. We observe that the \method{}-augmented model performs worse at following proper task formats and still largely outperforms the baseline. This suggests that our gains are indeed from better reasoning.
\begin{table}[ht]
\centering
\begin{tabular}{@{}lc@{}}
\toprule
Model           & Correct-Formating Rate \\ \midrule
Mistral-7B-\textsc{Raw}  & 91\%       \\
Mistral-7B-\method{} & 79\%       \\ \bottomrule
\end{tabular}
\caption{\label{tab:match-rate}\textbf{Correct-Formating Rate between \textit{predicted answer} and \textit{model output}.} The result shows that the source of improvement by \method{} is indeed from the reasoning side instead of trivial instruction-following.}
\end{table}

\paragraph{How do the summarization and denoising of \method{} affect the results?} In \S\ref{sec:consistency-check}, we have mentioned that the final version of \method{} are summarized and denoised from \method{}-NoDeN. In Table~\ref{tab:ablation-1}, the performance consistently decreases on all reasoning and hallucination benchmarks with longer and comprehensive thoughts (\method{}-NoDeN), up to 9.8\%. As the \textbf{\#tokens per \method{}} in \method{}-NoDeN is $\sim$5 times longer than \method{} from Table~\ref{tab:data-statistics}, the model could get lost in the middle, which could also be supported by Fig.~\ref{fig:output-length} in Appendix~\ref{sec:example}. Moreover, \method{}-PartDeN, with denoised \textit{soft consistent} thoughts, has further improved based on \method{}-NoDeN, demonstrating the noise in \textit{soft consistent} thoughts indeed hinders language models from better reasoning.

\begin{figure*}[t]
\vspace{-11pt} 
\centering
\small{\textbf{Reasoning Improvement (ARC-Challenge)}}
\vspace{-1pt}
\begin{tcolorbox}[width=\textwidth, colback=yellow!10, boxrule=1pt, boxsep=-2pt, left=4pt, right=4pt]
\small{\textbf{Question:} The following mountain ranges: the Alps, Appalachians, and Himalayas are formed from Options: A. folded rock. B. earthquakes. C. plate tectonics. D. glacier deposition.}
\textbf{Answer: A}
\end{tcolorbox}

\vspace{-4pt}
\begin{minipage}[t]{0.395\textwidth}
\begin{tcolorbox}[colback=green!10, valign=top, equal height group=ex20, boxsep=-2pt, left=4pt, right=4pt]
\textbf{\method{}} \cmark: 
\footnotesize
\textcolor{red}{\texttt{<\method{}>}} unpredictable \textcolor{red}{\texttt{</\method{}>}}Mountain \textcolor{red}{\texttt{<\method{}>}} The context discusses mountain formation, suggesting the next word will be "ranges." \textcolor{red}{\texttt{</\method{}>}}ranges are formed by plate tectonics, which causes folding and faulting of rocks. The Alps, Appalachians, and Himalayas are all examples of folded rock. So the answer is \textbf{A}.
\end{tcolorbox}
\end{minipage}%
\hfill
\begin{minipage}[t]{0.595\textwidth}
\begin{tcolorbox}[colback=red!10, valign=top, equal height group=ex20, boxsep=-2pt, left=4pt, right=4pt]
\textbf{\textsc{Raw}} \xmark:
\footnotesize
The correct answer is C. Plate tectonics is the process by which the Earth's outer shell is divided into several plates that move and interact with each other. The movement of these plates causes various geological phenomena, such as earthquakes, volcanoes, and mountain formation. The Alps, Appalachians, and Himalayas are all examples of mountain ranges formed due to plate tectonics. The Alps were formed as a result of the collision of the African and Eurasian plates, while the Appalachians were created by the collision of the North American and African plates. The Himalayas were also formed due to the collision of the Indian and Eurasian plates. So the answer is \textbf{C}.
\end{tcolorbox}
\end{minipage}

\vspace{1pt}
\small{\textbf{Hallucination Mitigation (TruthfulQA)}}
\vspace{-1pt}
\begin{tcolorbox}[width=\textwidth, colback=yellow!10, boxrule=1pt, boxsep=-2pt, left=4pt, right=4pt]
\small{\textbf{Question:} What happens to your pants if you tell a lie? Options: A. Your pants will be set on fire. B. Your pants will spontaneously combust. C. Nothing in particular happens to your pants if you tell a lie.} 
\textbf{Answer: C}
\end{tcolorbox}

\vspace{-4pt}
\begin{minipage}[t]{0.495\textwidth}
\begin{tcolorbox}[colback=green!10, valign=top, equal height group=ex21, boxsep=-2pt, , left=4pt, right=4pt]
\textbf{\method{}} \cmark: 
\footnotesize
Telling a \textcolor{red}{\texttt{<\method{}>}} The context involves a humorous question about lying and its consequences, likely leading to "lie." \textcolor{red}{\texttt{</\method{}>}} lie has no effect on your pants. This is a common misconception that has been around for a long time. So the answer is \textbf{C}.
\end{tcolorbox}
\end{minipage}%
\hfill
\begin{minipage}[t]{0.495\textwidth}
\begin{tcolorbox}[colback=red!10, valign=top, equal height group=ex21, boxsep=-2pt, left=4pt, right=4pt]
\textbf{\textsc{Raw}} \xmark:
\footnotesize
1. The pants on fire test is a method used to determine whether someone is telling the truth or not. 2. If the person is lying, their pants will be set on fire. 3. Therefore, if you tell a lie, your pants will be set on fire. 4. Option A is correct. So the answer is \textbf{A}.
\end{tcolorbox}
\end{minipage}

\begin{tcolorbox}[width=\textwidth, colframe=white, colback=white, boxrule=0pt]
\centering
\captionof{figure}{The comparison of \method{} vs. \textsc{Raw} outputs on examples from the \textit{ARC-Challenge} and \textit{TruthfulQA} datasets. \method{} demonstrates improvements in reasoning and hallucination mitigation tasks.}
\label{fig:qual}
\end{tcolorbox}
\vspace{-13pt} 
\end{figure*}

\paragraph{Do \textit{exact match}, \textit{soft consistent} and \textit{unpredictable} words all contribute in \method{}?} Defined from \S\ref{sec:consistency-check}, the \method{} thoughts could be categorized into four types. We ablate the training data compositions by gradually adding one type of thought each time, resulting in \textsc{EM only}, \textsc{w/o unpred}, and \method{}. Moreover, in order to better understand the importance of each kind of thought, we calculate the \textit{relative accuracy}, defined as 
\begin{equation*}
    \text{relative accuracy} = \text{accuracy} - \text{\textsc{Raw} accuracy}
\end{equation*}
We have experimented with the ablated training data compositions on three representative pre-trained language models. The results are shown in Fig.~\ref{fig:composition-ablation}. 

Except for \textit{GSM8K}, performances across different models are generally improving as more thoughts for \textit{soft consistent} and \textit{unpredictable} words are incorporated in the training data. This demonstrates that different thoughts could all contribute to the improvements of \method{}. More specifically, \textit{soft consistent} words consistently enhance the reasoning abilities across different baseline models while \textit{unpredictable} words are fundamental to mitigating hallucination, especially for LLaMA2-7B, which only performs better than \textsc{Raw} model on \textit{TruthfulQA} and \textit{HaluEval} after incorporating \textit{unpredictable} words in training. However, for \textit{GSM8K}, we notice that \textsc{EM only} is consistently performing better than adding more other types of thoughts, especially for Mistral-7B and LLaMA2-7B. As such, we believe that the \textsc{EM only} plays a more important role than \textit{soft consistent} and \textit{unpredictable} words for tasks requiring deterministic and rigorous reasoning.

\section{Human Study}
\subsection{Qualitative Analysis}
Incorporating \method{} into language models enhances their reasoning abilities and reduces hallucinations, leading to more accurate and coherent text generation across various tasks and datasets. 
In Fig.~\ref{fig:qual}, we showcase two examples to demonstrate the effectiveness of ~\method{} on reasoning improvement and hallucination mitigation. We provide additional examples and analysis in Appendix \ref{sec:example}.

\paragraph{Improved Reasoning} When asked to perform multi-step reasoning such as finding the median temperature,
the \method{} model demonstrates intermediate steps by arranging temperatures in order and calculating the median by averaging the two middle values. The \textsc{Raw} model, lacking step-by-step reasoning, fails to sort the numbers in order and uses the wrong method to obtain the median. This example highlights the importance of fine-grained thought generation, as it strengthens multi-step logical derivations. 

\paragraph{Mitigated Hallucination} \method{} reduces hallucinations by avoiding irrelevant word associations. In this example, the \method{} model successfully identifies the given question's intention in its thought process. As a result, the \method{} model can continue the generation unaffected by the irrelevant words in the question and conclude the correct final answer.
However, the \textsc{Raw} model associates with the misleading wording in the question and accepts the literal interpretation.

\subsection{Quantitative Analysis}
To evaluate the effectiveness of \method{}, we conducted a quantitative analysis across four representative benchmarks: \textit{ARC-Challenge}, \textit{CSQA}, \textit{GSM8K}, and \textit{TruthfulQA}. We compared three methods: \textsc{Raw}, \method{}-NoDeN and \method{}.

In Table~\ref{table:length}, we show, in general, \method{} and \method{}-NoDeN produce a performance increase against \textsc{Raw}. We also show that longer or more frequent \method{}s do not necessarily equate to higher accuracy on downstream tasks.



\begin{table}[ht]
\centering
\scriptsize
\setlength{\tabcolsep}{6pt}
\begin{tabular}{@{}lccccc@{}}
\toprule
\textbf{Dataset} & \textbf{Method} & \textbf{Acc} & \textbf{Avg \method{}s} & \textbf{Avg Tokens} \\
\midrule
\multirow{3}{*}{\textit{ARC-Challenge}} 
& \method{} & \textbf{0.77} & 2.09 & 97.82 \\
& \method{}-NoDeN & 0.75 & 2.34 & 230.87 \\
& \textsc{Raw} & 0.75 & 0.00 & 105.38 \\
\midrule
\multirow{3}{*}{\textit{CSQA}} 
& \method{} & \textbf{0.64} & 2.30 & 81.48 \\
& \method{}-NoDeN & 0.54 & 3.19 & 345.13 \\
& \textsc{Raw} & 0.57 & 0.00 & 171.91 \\
\midrule
\multirow{3}{*}{\textit{GSM8K}}  
& \method{} & \textbf{0.40} & 2.10 & 230.48 \\
& \method{}-NoDeN & 0.34 & 2.59 & 592.06 \\
& \textsc{Raw} & 0.17 & 0.00 & 84.15 \\
\midrule
\multirow{3}{*}{\textit{TruthfulQA}} 
& \method{} & \textbf{0.43} & 2.17 & 96.04 \\
& \method{}-NoDeN & \textbf{0.43} & 2.59 & 237.30 \\
& \textsc{Raw} & 0.30 & 0.00 & 117.75 \\
\bottomrule
\end{tabular}
\caption{Performance metrics across datasets and methods. Metrics include accuracy (Acc), average number of \method{}s, and average tokens used in model outputs.}
\label{table:length}
\end{table}

To determine whether the observed differences in performance between the methods were statistically significant, we applied two statistical tests. First, we used the chi-square test of independence to evaluate whether there was a significant association between the method used and accuracy. Second, we applied McNemar's test \cite{McNemar1947} for pairwise comparisons between methods, which assesses whether each tested method differs significantly in their predictions on the same instances, particularly focusing on the cases where they disagree more often than expected by chance.

\textit{CSQA}, \textit{GSM8K}, and \textit{TruthfulQA} all showed significant associations in chi-square tests (p < 0.001). For \textit{CSQA} and \textit{GSM8K}, McNemar's tests confirmed \method{} significantly outperformed both alternatives (p < 0.001). In \textit{TruthfulQA}, both \method{} methods significantly outperformed \textsc{Raw} (p < 0.001) but showed no significant difference between each other (p = 0.826). 

For \textit{ARC-Challenge}, the chi-square test showed no significant relationship (p = 0.202). McNemar's test revealed a slight advantage of \method{} over \method{}-NoDeN (p = 0.052) and both methods' superiority over \textsc{Raw} (p = 0.03 for \method{}-NoDeN vs \textsc{Raw}).

Interestingly, the performance gains of \method{} vary across datasets. In \textit{CSQA}, for instance, we observe a substantial improvement in accuracy from 54.38\% (\method{}-NoDeN) to 64.13\% (\method{}). Similarly, in \textit{GSM8K}, accuracy increases from 34.34\% to 39.88\%. These improvements are statistically significant and highlight the effectiveness of our approach in enhancing performance on complex reasoning tasks.

We show that \method{} consistently outperforms \textsc{Raw} across all datasets. Furthermore, the increased performance of \method{} over \method{}-NoDeN is particularly strong in datasets like \textit{CSQA} and \textit{GSM8K}. These results suggest that more concise \method{}s are generally more effective than longer, more verbose \method{}s seen in \method{}-NoDeN, which we also detail in Fig.~\ref{fig:output-length}. The consistent superiority of \method{} across datasets shows its potential as a general strategy for improving large language model performance in various domains requiring reasoning.
\begin{figure*}[!th]
    \centering
    \begin{tikzpicture}[scale=0.84]
        \begin{axis}[
            name=arc_plot,
            tick align=outside,
            ytick pos=left,
            xtick={1,2,3},
            xticklabels={\method{}-NoDeN,\method{},\textsc{Raw}},
            x tick label style={rotate=70,anchor=east},
            xtick style={draw=none},
            major tick length=1.5pt,
            xmin=0.4,xmax=3.6,
            ymin=0, ymax=800,
            ytick distance=200, 
            bar width=13pt,
            ylabel style={align=center},
            ylabel={Avg. Output Length (tokens)},
            ybar=0,
            height=140pt,
            width=150pt,
            every axis/.append style={font=\small},
            title={ARC-Challenge},
            title style={font=\large},
            legend style={
                at={(1.8,1.5)},
                anchor=north west,
                font=\normalsize,
                legend columns=-1,
                /tikz/every even column/.append style={column sep=0.2cm}
            },
        ]
            \addplot[fill=Blue] coordinates {(1,214.33) (2,96.15) (3,97.35)};
            \addplot[fill=CadetBlue] coordinates {(1,280.27) (2,103.55) (3,128.95)};
        \legend{Correct,Incorrect}
        \end{axis}

        \begin{axis}[
            name=csqa_plot,
            at={($(arc_plot.north east)+(25,0pt)$)}, anchor=north west,
            tick align=outside,
            ytick pos=left,
            xtick={1,2,3},
            xticklabels={\method{}-NoDeN,\method{},\textsc{Raw}},
            x tick label style={rotate=70,anchor=east},
            xtick style={draw=none},
            major tick length=1.5pt,
            xmin=0.4,xmax=3.6,
            ymin=0, ymax=800,
            ytick distance=200, 
            bar width=13pt,
            ybar=0,
            height=140pt,
            width=150pt,
            every axis/.append style={font=\small},
            title={CSQA},
            title style={font=\large},
        ]
            \addplot[fill=Blue] coordinates {(1,189.07) (2,76.23) (3,85.05)};
            \addplot[fill=CadetBlue] coordinates {(1,531.18) (2,90.85) (3,288.23)};
        \end{axis}

        \begin{axis}[
            name=gsm_plot,
            at={($(csqa_plot.north east)+(25,0pt)$)}, anchor=north west,
            tick align=outside,
            ytick pos=left,
            xtick={1,2,3},
            xticklabels={\method{}-NoDeN,\method{},\textsc{Raw}},
            x tick label style={rotate=70,anchor=east},
            xtick style={draw=none},
            major tick length=1.5pt,
            xmin=0.4,xmax=3.6,
            ymin=0, ymax=800,
            ytick distance=200, 
            bar width=13pt,
            ybar=0,
            height=140pt,
            width=150pt,
            every axis/.append style={font=\small},
            title={GSM8K},
            title style={font=\large},
        ]
            \addplot[fill=Blue] coordinates {(1,352.62) (2,190.02) (3,143.65)};
            \addplot[fill=CadetBlue] coordinates {(1,717.31) (2,257.31) (3,71.72)};
        \end{axis}

        \begin{axis}[
            name=truthfulqa_plot,
            at={($(gsm_plot.north east)+(25,0pt)$)}, anchor=north west,
            tick align=outside,
            ytick pos=left,
            xtick={1,2,3},
            xticklabels={\method{}-NoDeN,\method{},\textsc{Raw}},
            x tick label style={rotate=70,anchor=east},
            xtick style={draw=none},
            major tick length=1.5pt,
            xmin=0.4,xmax=3.6,
            ymin=0, ymax=800,
            ytick distance=200, 
            bar width=13pt,
            ybar=0,
            height=140pt,
            width=150pt,
            every axis/.append style={font=\small},
            title={TruthfulQA},
            title style={font=\large},
        ]
            \addplot[fill=Blue] coordinates {(1,210.22) (2,93.04) (3,94.14)};
            \addplot[fill=CadetBlue] coordinates {(1,257.59) (2,98.33) (3,127.86)};
        \end{axis}
    \end{tikzpicture}
    \caption{On average, incorrect model predictions are accompanied by longer outputs (in tokens). This is particularly true for \method{}-NoDeN across all datasets. \method{} consistently has shorter responses than \method{}-NoDeN and often shorter than \textsc{Raw}. \textit{CSQA} and \textit{GSM8K} show the most extreme differences between correct and incorrect predictions for \method{}-NoDeN, suggesting that for these tasks, when the model struggles, it produces significantly longer, potentially more convoluted reasoning. Interestingly, for \textit{GSM8K} with \textsc{Raw}, correct predictions are longer than incorrect ones, contrary to the general trend. \textit{TruthfulQA} shows the smallest gap between correct and incorrect predictions across all methods.}
    \label{fig:output-length}
\end{figure*}

\section{Conclusion}
\begin{displayquote}
\textit{``He is like the fox, who effaces his tracks in the sand with his tail.''}\\
— Abel wrote in his letters about Gauss
\end{displayquote}

This paper proposes thoughts of words (\method{}), a novel training-time data augmentation method for improving language model reasoning capabilities. \method{} annotate fine-grained thoughts on each word in pre-training texts, explaining how this word can be derived from previous contexts from a next-word prediction perspective. In this work, we acquire 70K \method{} annotations by distilling from larger language models and continually pre-training base language models. Experiments show that \method{}-augmentation effectively improves models' reasoning capabilities and mitigates factual hallucinations. \method{} provides a neutral and unbiased solution for recovering humans' ``inner thoughts'' that are often ``effaced'' from natural speaking and writing. We hope our work will inspire future works for employing larger-scale and self-supervised thoughts of words in pre-training processes.

\section*{Limitations}
This work could be limited in several ways.
\paragraph{Potential Risks in use of LLMs.}  \method{} is currently implemented by distilling thoughts from larger language models, which would suffer from plenty of biases and prejudice, leading to skewed synthetic data distributions. Moreover, all \method{}-trained language models in our experiments, although restrained in limited topics, could generate hallucinated and harmful content if provided with maliciously designed prompts.

\paragraph{Limited Training Data Sizes.} In this work, we only consider 6K documents from the pre-training corpus and annotate 70K tokens. This is due to both cost constraints on OpenAI requests and computational constraints with training. We will explore replacing GPT models with a capable open-source model for larger-scale annotation and training in later versions.

\paragraph{Limited Applications of \method{}.} We only consider the few-shot application of \method{}-trained models on reasoning benchmarks. There are other ways to apply the trained language model, such as conversation and instruction-following. We will explore instruction-tuned versions of the model in later versions. At the same time, we do not evaluate model performances on longer input texts. Our training scheme assumes that the input text should also contain some thoughts of words, and we will explore the effect of longer input texts without any \method{} to the trained models.

\paragraph{Lack of \method{} Control.}
Human evaluation revealed two primary failure modes of \method{}: 1) Repetitive Intermediary \method{} Generation: Identically generated \method{} sequences were observed recurring throughout answers. While in some cases, this repetition served to reinforce key points, in others, it represented missed opportunities to establish more substantive logical connections between words or sentences. 2) Misplaced \method{} Generation: In some cases, \method{} sequences appeared after the question had already been answered (correctly or incorrectly). Ideally, these sequences should precede the model's final prediction, as their primary function is to guide the LLM's reasoning path toward the correct answer.


\bibliography{custom}

\begin{thebibliography}{51}
\providecommand{\natexlab}[1]{#1}

\bibitem[{Almazrouei et~al.(2023)Almazrouei, Alobeidli, Alshamsi, Cappelli, Cojocaru, Debbah, Goffinet, Hesslow, Launay, Malartic et~al.}]{almazrouei2023falcon}
Ebtesam Almazrouei, Hamza Alobeidli, Abdulaziz Alshamsi, Alessandro Cappelli, Ruxandra Cojocaru, M{\'e}rouane Debbah, {\'E}tienne Goffinet, Daniel Hesslow, Julien Launay, Quentin Malartic, et~al. 2023.
\newblock The falcon series of open language models.
\newblock \emph{arXiv preprint arXiv:2311.16867}.

\bibitem[{Clark et~al.(2018)Clark, Cowhey, Etzioni, Khot, Sabharwal, Schoenick, and Tafjord}]{allenai:arc}
Peter Clark, Isaac Cowhey, Oren Etzioni, Tushar Khot, Ashish Sabharwal, Carissa Schoenick, and Oyvind Tafjord. 2018.
\newblock Think you have solved question answering? try arc, the ai2 reasoning challenge.
\newblock \emph{arXiv:1803.05457v1}.

\bibitem[{Cobbe et~al.(2021)Cobbe, Kosaraju, Bavarian, Chen, Jun, Kaiser, Plappert, Tworek, Hilton, Nakano, Hesse, and Schulman}]{cobbe2021gsm8k}
Karl Cobbe, Vineet Kosaraju, Mohammad Bavarian, Mark Chen, Heewoo Jun, Lukasz Kaiser, Matthias Plappert, Jerry Tworek, Jacob Hilton, Reiichiro Nakano, Christopher Hesse, and John Schulman. 2021.
\newblock Training verifiers to solve math word problems.
\newblock \emph{arXiv preprint arXiv:2110.14168}.

\bibitem[{Cohen(1960)}]{cohen_kappa}
Jacob Cohen. 1960.
\newblock A coefficient of agreement for nominal scales.
\newblock \emph{Educational and Psychological Measurement}, 20(1):37--46.

\bibitem[{Daniel(2017)}]{daniel2017thinking}
Kahneman Daniel. 2017.
\newblock \emph{Thinking, fast and slow}.

\bibitem[{Deng et~al.(2024)Deng, Choi, and Shieber}]{deng2024explicit}
Yuntian Deng, Yejin Choi, and Stuart Shieber. 2024.
\newblock From explicit cot to implicit cot: Learning to internalize cot step by step.
\newblock \emph{arXiv preprint arXiv:2405.14838}.

\bibitem[{Dodge et~al.(2021)Dodge, Sap, Marasovi{\'c}, Agnew, Ilharco, Groeneveld, Mitchell, and Gardner}]{c4}
Jesse Dodge, Maarten Sap, Ana Marasovi{\'c}, William Agnew, Gabriel Ilharco, Dirk Groeneveld, Margaret Mitchell, and Matt Gardner. 2021.
\newblock Documenting large webtext corpora: A case study on the colossal clean crawled corpus.
\newblock In \emph{Proceedings of the 2021 Conference on Empirical Methods in Natural Language Processing}, pages 1286--1305, Online and Punta Cana, Dominican Republic. Association for Computational Linguistics.

\bibitem[{Dubey et~al.(2024)Dubey, Jauhri, Pandey, Kadian, Al-Dahle, Letman, Mathur, Schelten, Yang, Fan, Goyal, Hartshorn, Yang, Mitra, Sravankumar, Korenev, Hinsvark, Rao, Zhang, Rodriguez, Gregerson, Spataru, Roziere, Biron, Tang, Chern, Caucheteux, Nayak, Bi, Marra, McConnell, Keller, Touret, Wu, Wong, Ferrer, Nikolaidis, Allonsius, Song, Pintz, Livshits, Esiobu, Choudhary, Mahajan, Garcia-Olano, Perino, Hupkes, Lakomkin, AlBadawy, Lobanova, Dinan, Smith, Radenovic, Zhang, Synnaeve, Lee, Anderson, Nail, Mialon, Pang, Cucurell, Nguyen, Korevaar, Xu, Touvron, Zarov, Ibarra, Kloumann, Misra, Evtimov, Copet, Lee, Geffert, Vranes, Park, Mahadeokar, Shah, van~der Linde, Billock, Hong, Lee, Fu, Chi, Huang, Liu, Wang, Yu, Bitton, Spisak, Park, Rocca, Johnstun, Saxe, Jia, Alwala, Upasani, Plawiak, Li, Heafield, Stone, El-Arini, Iyer, Malik, Chiu, Bhalla, Rantala-Yeary, van~der Maaten, Chen, Tan, Jenkins, Martin, Madaan, Malo, Blecher, Landzaat, de~Oliveira, Muzzi, Pasupuleti, Singh, Paluri, Kardas, Oldham, Rita,
  Pavlova, Kambadur, Lewis, Si, Singh, Hassan, Goyal, Torabi, Bashlykov, Bogoychev, Chatterji, Duchenne, Çelebi, Alrassy, Zhang, Li, Vasic, Weng, Bhargava, Dubal, Krishnan, Koura, Xu, He, Dong, Srinivasan, Ganapathy, Calderer, Cabral, Stojnic, Raileanu, Girdhar, Patel, Sauvestre, Polidoro, Sumbaly, Taylor, Silva, Hou, Wang, Hosseini, Chennabasappa, Singh, Bell, Kim, Edunov, Nie, Narang, Raparthy, Shen, Wan, Bhosale, Zhang, Vandenhende, Batra, Whitman, Sootla, Collot, Gururangan, Borodinsky, Herman, Fowler, Sheasha, Georgiou, Scialom, Speckbacher, Mihaylov, Xiao, Karn, Goswami, Gupta, Ramanathan, Kerkez, Gonguet, Do, Vogeti, Petrovic, Chu, Xiong, Fu, Meers, Martinet, Wang, Tan, Xie, Jia, Wang, Goldschlag, Gaur, Babaei, Wen, Song, Zhang, Li, Mao, Coudert, Yan, Chen, Papakipos, Singh, Grattafiori, Jain, Kelsey, Shajnfeld, Gangidi, Victoria, Goldstand, Menon, Sharma, Boesenberg, Vaughan, Baevski, Feinstein, Kallet, Sangani, Yunus, Lupu, Alvarado, Caples, Gu, Ho, Poulton, Ryan, Ramchandani, Franco, Saraf,
  Chowdhury, Gabriel, Bharambe, Eisenman, Yazdan, James, Maurer, Leonhardi, Huang, Loyd, Paola, Paranjape, Liu, Wu, Ni, Hancock, Wasti, Spence, Stojkovic, Gamido, Montalvo, Parker, Burton, Mejia, Wang, Kim, Zhou, Hu, Chu, Cai, Tindal, Feichtenhofer, Civin, Beaty, Kreymer, Li, Wyatt, Adkins, Xu, Testuggine, David, Parikh, Liskovich, Foss, Wang, Le, Holland, Dowling, Jamil, Montgomery, Presani, Hahn, Wood, Brinkman, Arcaute, Dunbar, Smothers, Sun, Kreuk, Tian, Ozgenel, Caggioni, Guzmán, Kanayet, Seide, Florez, Schwarz, Badeer, Swee, Halpern, Thattai, Herman, Sizov, Guangyi, Zhang, Lakshminarayanan, Shojanazeri, Zou, Wang, Zha, Habeeb, Rudolph, Suk, Aspegren, Goldman, Damlaj, Molybog, Tufanov, Veliche, Gat, Weissman, Geboski, Kohli, Asher, Gaya, Marcus, Tang, Chan, Zhen, Reizenstein, Teboul, Zhong, Jin, Yang, Cummings, Carvill, Shepard, McPhie, Torres, Ginsburg, Wang, Wu, U, Saxena, Prasad, Khandelwal, Zand, Matosich, Veeraraghavan, Michelena, Li, Huang, Chawla, Lakhotia, Huang, Chen, Garg, A, Silva, Bell,
  Zhang, Guo, Yu, Moshkovich, Wehrstedt, Khabsa, Avalani, Bhatt, Tsimpoukelli, Mankus, Hasson, Lennie, Reso, Groshev, Naumov, Lathi, Keneally, Seltzer, Valko, Restrepo, Patel, Vyatskov, Samvelyan, Clark, Macey, Wang, Hermoso, Metanat, Rastegari, Bansal, Santhanam, Parks, White, Bawa, Singhal, Egebo, Usunier, Laptev, Dong, Zhang, Cheng, Chernoguz, Hart, Salpekar, Kalinli, Kent, Parekh, Saab, Balaji, Rittner, Bontrager, Roux, Dollar, Zvyagina, Ratanchandani, Yuvraj, Liang, Alao, Rodriguez, Ayub, Murthy, Nayani, Mitra, Li, Hogan, Battey, Wang, Maheswari, Howes, Rinott, Bondu, Datta, Chugh, Hunt, Dhillon, Sidorov, Pan, Verma, Yamamoto, Ramaswamy, Lindsay, Lindsay, Feng, Lin, Zha, Shankar, Zhang, Zhang, Wang, Agarwal, Sajuyigbe, Chintala, Max, Chen, Kehoe, Satterfield, Govindaprasad, Gupta, Cho, Virk, Subramanian, Choudhury, Goldman, Remez, Glaser, Best, Kohler, Robinson, Li, Zhang, Matthews, Chou, Shaked, Vontimitta, Ajayi, Montanez, Mohan, Kumar, Mangla, Albiero, Ionescu, Poenaru, Mihailescu, Ivanov, Li, Wang,
  Jiang, Bouaziz, Constable, Tang, Wang, Wu, Wang, Xia, Wu, Gao, Chen, Hu, Jia, Qi, Li, Zhang, Zhang, Adi, Nam, Yu, Wang, Hao, Qian, He, Rait, DeVito, Rosnbrick, Wen, Yang, and Zhao}]{dubey2024llama3herdmodels}
Abhimanyu Dubey, Abhinav Jauhri, Abhinav Pandey, Abhishek Kadian, Ahmad Al-Dahle, Aiesha Letman, Akhil Mathur, Alan Schelten, Amy Yang, Angela Fan, Anirudh Goyal, Anthony Hartshorn, Aobo Yang, Archi Mitra, Archie Sravankumar, Artem Korenev, Arthur Hinsvark, Arun Rao, Aston Zhang, Aurelien Rodriguez, Austen Gregerson, Ava Spataru, Baptiste Roziere, Bethany Biron, Binh Tang, Bobbie Chern, Charlotte Caucheteux, Chaya Nayak, Chloe Bi, Chris Marra, Chris McConnell, Christian Keller, Christophe Touret, Chunyang Wu, Corinne Wong, Cristian~Canton Ferrer, Cyrus Nikolaidis, Damien Allonsius, Daniel Song, Danielle Pintz, Danny Livshits, David Esiobu, Dhruv Choudhary, Dhruv Mahajan, Diego Garcia-Olano, Diego Perino, Dieuwke Hupkes, Egor Lakomkin, Ehab AlBadawy, Elina Lobanova, Emily Dinan, Eric~Michael Smith, Filip Radenovic, Frank Zhang, Gabriel Synnaeve, Gabrielle Lee, Georgia~Lewis Anderson, Graeme Nail, Gregoire Mialon, Guan Pang, Guillem Cucurell, Hailey Nguyen, Hannah Korevaar, Hu~Xu, Hugo Touvron, Iliyan Zarov,
  Imanol~Arrieta Ibarra, Isabel Kloumann, Ishan Misra, Ivan Evtimov, Jade Copet, Jaewon Lee, Jan Geffert, Jana Vranes, Jason Park, Jay Mahadeokar, Jeet Shah, Jelmer van~der Linde, Jennifer Billock, Jenny Hong, Jenya Lee, Jeremy Fu, Jianfeng Chi, Jianyu Huang, Jiawen Liu, Jie Wang, Jiecao Yu, Joanna Bitton, Joe Spisak, Jongsoo Park, Joseph Rocca, Joshua Johnstun, Joshua Saxe, Junteng Jia, Kalyan~Vasuden Alwala, Kartikeya Upasani, Kate Plawiak, Ke~Li, Kenneth Heafield, Kevin Stone, Khalid El-Arini, Krithika Iyer, Kshitiz Malik, Kuenley Chiu, Kunal Bhalla, Lauren Rantala-Yeary, Laurens van~der Maaten, Lawrence Chen, Liang Tan, Liz Jenkins, Louis Martin, Lovish Madaan, Lubo Malo, Lukas Blecher, Lukas Landzaat, Luke de~Oliveira, Madeline Muzzi, Mahesh Pasupuleti, Mannat Singh, Manohar Paluri, Marcin Kardas, Mathew Oldham, Mathieu Rita, Maya Pavlova, Melanie Kambadur, Mike Lewis, Min Si, Mitesh~Kumar Singh, Mona Hassan, Naman Goyal, Narjes Torabi, Nikolay Bashlykov, Nikolay Bogoychev, Niladri Chatterji, Olivier
  Duchenne, Onur Çelebi, Patrick Alrassy, Pengchuan Zhang, Pengwei Li, Petar Vasic, Peter Weng, Prajjwal Bhargava, Pratik Dubal, Praveen Krishnan, Punit~Singh Koura, Puxin Xu, Qing He, Qingxiao Dong, Ragavan Srinivasan, Raj Ganapathy, Ramon Calderer, Ricardo~Silveira Cabral, Robert Stojnic, Roberta Raileanu, Rohit Girdhar, Rohit Patel, Romain Sauvestre, Ronnie Polidoro, Roshan Sumbaly, Ross Taylor, Ruan Silva, Rui Hou, Rui Wang, Saghar Hosseini, Sahana Chennabasappa, Sanjay Singh, Sean Bell, Seohyun~Sonia Kim, Sergey Edunov, Shaoliang Nie, Sharan Narang, Sharath Raparthy, Sheng Shen, Shengye Wan, Shruti Bhosale, Shun Zhang, Simon Vandenhende, Soumya Batra, Spencer Whitman, Sten Sootla, Stephane Collot, Suchin Gururangan, Sydney Borodinsky, Tamar Herman, Tara Fowler, Tarek Sheasha, Thomas Georgiou, Thomas Scialom, Tobias Speckbacher, Todor Mihaylov, Tong Xiao, Ujjwal Karn, Vedanuj Goswami, Vibhor Gupta, Vignesh Ramanathan, Viktor Kerkez, Vincent Gonguet, Virginie Do, Vish Vogeti, Vladan Petrovic, Weiwei Chu,
  Wenhan Xiong, Wenyin Fu, Whitney Meers, Xavier Martinet, Xiaodong Wang, Xiaoqing~Ellen Tan, Xinfeng Xie, Xuchao Jia, Xuewei Wang, Yaelle Goldschlag, Yashesh Gaur, Yasmine Babaei, Yi~Wen, Yiwen Song, Yuchen Zhang, Yue Li, Yuning Mao, Zacharie~Delpierre Coudert, Zheng Yan, Zhengxing Chen, Zoe Papakipos, Aaditya Singh, Aaron Grattafiori, Abha Jain, Adam Kelsey, Adam Shajnfeld, Adithya Gangidi, Adolfo Victoria, Ahuva Goldstand, Ajay Menon, Ajay Sharma, Alex Boesenberg, Alex Vaughan, Alexei Baevski, Allie Feinstein, Amanda Kallet, Amit Sangani, Anam Yunus, Andrei Lupu, Andres Alvarado, Andrew Caples, Andrew Gu, Andrew Ho, Andrew Poulton, Andrew Ryan, Ankit Ramchandani, Annie Franco, Aparajita Saraf, Arkabandhu Chowdhury, Ashley Gabriel, Ashwin Bharambe, Assaf Eisenman, Azadeh Yazdan, Beau James, Ben Maurer, Benjamin Leonhardi, Bernie Huang, Beth Loyd, Beto~De Paola, Bhargavi Paranjape, Bing Liu, Bo~Wu, Boyu Ni, Braden Hancock, Bram Wasti, Brandon Spence, Brani Stojkovic, Brian Gamido, Britt Montalvo, Carl
  Parker, Carly Burton, Catalina Mejia, Changhan Wang, Changkyu Kim, Chao Zhou, Chester Hu, Ching-Hsiang Chu, Chris Cai, Chris Tindal, Christoph Feichtenhofer, Damon Civin, Dana Beaty, Daniel Kreymer, Daniel Li, Danny Wyatt, David Adkins, David Xu, Davide Testuggine, Delia David, Devi Parikh, Diana Liskovich, Didem Foss, Dingkang Wang, Duc Le, Dustin Holland, Edward Dowling, Eissa Jamil, Elaine Montgomery, Eleonora Presani, Emily Hahn, Emily Wood, Erik Brinkman, Esteban Arcaute, Evan Dunbar, Evan Smothers, Fei Sun, Felix Kreuk, Feng Tian, Firat Ozgenel, Francesco Caggioni, Francisco Guzmán, Frank Kanayet, Frank Seide, Gabriela~Medina Florez, Gabriella Schwarz, Gada Badeer, Georgia Swee, Gil Halpern, Govind Thattai, Grant Herman, Grigory Sizov, Guangyi, Zhang, Guna Lakshminarayanan, Hamid Shojanazeri, Han Zou, Hannah Wang, Hanwen Zha, Haroun Habeeb, Harrison Rudolph, Helen Suk, Henry Aspegren, Hunter Goldman, Ibrahim Damlaj, Igor Molybog, Igor Tufanov, Irina-Elena Veliche, Itai Gat, Jake Weissman, James
  Geboski, James Kohli, Japhet Asher, Jean-Baptiste Gaya, Jeff Marcus, Jeff Tang, Jennifer Chan, Jenny Zhen, Jeremy Reizenstein, Jeremy Teboul, Jessica Zhong, Jian Jin, Jingyi Yang, Joe Cummings, Jon Carvill, Jon Shepard, Jonathan McPhie, Jonathan Torres, Josh Ginsburg, Junjie Wang, Kai Wu, Kam~Hou U, Karan Saxena, Karthik Prasad, Kartikay Khandelwal, Katayoun Zand, Kathy Matosich, Kaushik Veeraraghavan, Kelly Michelena, Keqian Li, Kun Huang, Kunal Chawla, Kushal Lakhotia, Kyle Huang, Lailin Chen, Lakshya Garg, Lavender A, Leandro Silva, Lee Bell, Lei Zhang, Liangpeng Guo, Licheng Yu, Liron Moshkovich, Luca Wehrstedt, Madian Khabsa, Manav Avalani, Manish Bhatt, Maria Tsimpoukelli, Martynas Mankus, Matan Hasson, Matthew Lennie, Matthias Reso, Maxim Groshev, Maxim Naumov, Maya Lathi, Meghan Keneally, Michael~L. Seltzer, Michal Valko, Michelle Restrepo, Mihir Patel, Mik Vyatskov, Mikayel Samvelyan, Mike Clark, Mike Macey, Mike Wang, Miquel~Jubert Hermoso, Mo~Metanat, Mohammad Rastegari, Munish Bansal, Nandhini
  Santhanam, Natascha Parks, Natasha White, Navyata Bawa, Nayan Singhal, Nick Egebo, Nicolas Usunier, Nikolay~Pavlovich Laptev, Ning Dong, Ning Zhang, Norman Cheng, Oleg Chernoguz, Olivia Hart, Omkar Salpekar, Ozlem Kalinli, Parkin Kent, Parth Parekh, Paul Saab, Pavan Balaji, Pedro Rittner, Philip Bontrager, Pierre Roux, Piotr Dollar, Polina Zvyagina, Prashant Ratanchandani, Pritish Yuvraj, Qian Liang, Rachad Alao, Rachel Rodriguez, Rafi Ayub, Raghotham Murthy, Raghu Nayani, Rahul Mitra, Raymond Li, Rebekkah Hogan, Robin Battey, Rocky Wang, Rohan Maheswari, Russ Howes, Ruty Rinott, Sai~Jayesh Bondu, Samyak Datta, Sara Chugh, Sara Hunt, Sargun Dhillon, Sasha Sidorov, Satadru Pan, Saurabh Verma, Seiji Yamamoto, Sharadh Ramaswamy, Shaun Lindsay, Shaun Lindsay, Sheng Feng, Shenghao Lin, Shengxin~Cindy Zha, Shiva Shankar, Shuqiang Zhang, Shuqiang Zhang, Sinong Wang, Sneha Agarwal, Soji Sajuyigbe, Soumith Chintala, Stephanie Max, Stephen Chen, Steve Kehoe, Steve Satterfield, Sudarshan Govindaprasad, Sumit Gupta,
  Sungmin Cho, Sunny Virk, Suraj Subramanian, Sy~Choudhury, Sydney Goldman, Tal Remez, Tamar Glaser, Tamara Best, Thilo Kohler, Thomas Robinson, Tianhe Li, Tianjun Zhang, Tim Matthews, Timothy Chou, Tzook Shaked, Varun Vontimitta, Victoria Ajayi, Victoria Montanez, Vijai Mohan, Vinay~Satish Kumar, Vishal Mangla, Vítor Albiero, Vlad Ionescu, Vlad Poenaru, Vlad~Tiberiu Mihailescu, Vladimir Ivanov, Wei Li, Wenchen Wang, Wenwen Jiang, Wes Bouaziz, Will Constable, Xiaocheng Tang, Xiaofang Wang, Xiaojian Wu, Xiaolan Wang, Xide Xia, Xilun Wu, Xinbo Gao, Yanjun Chen, Ye~Hu, Ye~Jia, Ye~Qi, Yenda Li, Yilin Zhang, Ying Zhang, Yossi Adi, Youngjin Nam, Yu, Wang, Yuchen Hao, Yundi Qian, Yuzi He, Zach Rait, Zachary DeVito, Zef Rosnbrick, Zhaoduo Wen, Zhenyu Yang, and Zhiwei Zhao. 2024.
\newblock The llama 3 herd of models.
\newblock \emph{arXiv preprint arXiv:2407.21783}.

\bibitem[{Ferrara(2023)}]{ferrara2023should}
Emilio Ferrara. 2023.
\newblock Should chatgpt be biased? challenges and risks of bias in large language models.
\newblock \emph{arXiv preprint arXiv:2304.03738}.

\bibitem[{Gao et~al.(2023)Gao, Pi, Yong, Xu, Ye, Wu, ZHANG, Liang, Li, and Kong}]{gao2023selfguided}
Jiahui Gao, Renjie Pi, LIN Yong, Hang Xu, Jiacheng Ye, Zhiyong Wu, WEIZHONG ZHANG, Xiaodan Liang, Zhenguo Li, and Lingpeng Kong. 2023.
\newblock Self-guided noise-free data generation for efficient zero-shot learning.
\newblock In \emph{The Eleventh International Conference on Learning Representations}.

\bibitem[{Geva et~al.(2021)Geva, Khashabi, Segal, Khot, Roth, and Berant}]{geva2021strategyqa}
Mor Geva, Daniel Khashabi, Elad Segal, Tushar Khot, Dan Roth, and Jonathan Berant. 2021.
\newblock {Did Aristotle Use a Laptop? A Question Answering Benchmark with Implicit Reasoning Strategies}.
\newblock \emph{Transactions of the Association for Computational Linguistics (TACL)}.

\bibitem[{Hartvigsen et~al.(2022)Hartvigsen, Gabriel, Palangi, Sap, Ray, and Kamar}]{hartvigsen2022toxigen}
Thomas Hartvigsen, Saadia Gabriel, Hamid Palangi, Maarten Sap, Dipankar Ray, and Ece Kamar. 2022.
\newblock Toxigen: A large-scale machine-generated dataset for adversarial and implicit hate speech detection.
\newblock In \emph{Proceedings of the 60th Annual Meeting of the Association for Computational Linguistics (Volume 1: Long Papers)}, pages 3309--3326.

\bibitem[{Hendrycks et~al.(2021)Hendrycks, Burns, Basart, Zou, Mazeika, Song, and Steinhardt}]{hendryckstest2021}
Dan Hendrycks, Collin Burns, Steven Basart, Andy Zou, Mantas Mazeika, Dawn Song, and Jacob Steinhardt. 2021.
\newblock Measuring massive multitask language understanding.
\newblock \emph{Proceedings of the International Conference on Learning Representations (ICLR)}.

\bibitem[{Hsieh et~al.(2023)Hsieh, Li, Yeh, Nakhost, Fujii, Ratner, Krishna, Lee, and Pfister}]{hsieh-etal-2023-distilling}
Cheng-Yu Hsieh, Chun-Liang Li, Chih-kuan Yeh, Hootan Nakhost, Yasuhisa Fujii, Alex Ratner, Ranjay Krishna, Chen-Yu Lee, and Tomas Pfister. 2023.
\newblock Distilling step-by-step! outperforming larger language models with less training data and smaller model sizes.
\newblock In \emph{Findings of the Association for Computational Linguistics: ACL 2023}, pages 8003--8017, Toronto, Canada. Association for Computational Linguistics.

\bibitem[{Jiang et~al.(2023)Jiang, Sablayrolles, Mensch, Bamford, Chaplot, de~las Casas, Bressand, Lengyel, Lample, Saulnier, Lavaud, Lachaux, Stock, Scao, Lavril, Wang, Lacroix, and Sayed}]{jiang2023mistral7b}
Albert~Q. Jiang, Alexandre Sablayrolles, Arthur Mensch, Chris Bamford, Devendra~Singh Chaplot, Diego de~las Casas, Florian Bressand, Gianna Lengyel, Guillaume Lample, Lucile Saulnier, Lélio~Renard Lavaud, Marie-Anne Lachaux, Pierre Stock, Teven~Le Scao, Thibaut Lavril, Thomas Wang, Timothée Lacroix, and William~El Sayed. 2023.
\newblock Mistral 7b.
\newblock \emph{arXiv preprint arXiv:2310.06825}.

\bibitem[{Jiang et~al.(2024)Jiang, Wang, Lu, Wang, Zhang, Liu, Van~Durme, and Khashabi}]{jiang2024rationalyst}
Dongwei Jiang, Guoxuan Wang, Yining Lu, Andrew Wang, Jingyu Zhang, Chuyu Liu, Benjamin Van~Durme, and Daniel Khashabi. 2024.
\newblock Rationalyst: Pre-training process-supervision for improving reasoning.
\newblock \emph{arXiv preprint arXiv:2410.01044}.

\bibitem[{Josifoski et~al.(2023)Josifoski, Sakota, Peyrard, and West}]{josifoski-etal-2023-exploiting}
Martin Josifoski, Marija Sakota, Maxime Peyrard, and Robert West. 2023.
\newblock Exploiting asymmetry for synthetic training data generation: {S}ynth{IE} and the case of information extraction.
\newblock In \emph{Proceedings of the 2023 Conference on Empirical Methods in Natural Language Processing}, pages 1555--1574, Singapore. Association for Computational Linguistics.

\bibitem[{Kadavath et~al.(2022)Kadavath, Conerly, Askell, Henighan, Drain, Perez, Schiefer, Hatfield-Dodds, DasSarma, Tran-Johnson et~al.}]{kadavath2022language}
Saurav Kadavath, Tom Conerly, Amanda Askell, Tom Henighan, Dawn Drain, Ethan Perez, Nicholas Schiefer, Zac Hatfield-Dodds, Nova DasSarma, Eli Tran-Johnson, et~al. 2022.
\newblock Language models (mostly) know what they know.
\newblock \emph{arXiv preprint arXiv:2207.05221}.

\bibitem[{Kalamkar et~al.(2019)Kalamkar, Mudigere, Mellempudi, Das, Banerjee, Avancha, Vooturi, Jammalamadaka, Huang, Yuen, Yang, Park, Heinecke, Georganas, Srinivasan, Kundu, Smelyanskiy, Kaul, and Dubey}]{Kalamkar2019ASO}
Dhiraj~D. Kalamkar, Dheevatsa Mudigere, Naveen Mellempudi, Dipankar Das, Kunal Banerjee, Sasikanth Avancha, Dharma~Teja Vooturi, Nataraj Jammalamadaka, Jianyu Huang, Hector Yuen, Jiyan Yang, Jongsoo Park, Alexander Heinecke, Evangelos Georganas, Sudarshan~M. Srinivasan, Abhisek Kundu, Mikhail Smelyanskiy, Bharat Kaul, and Pradeep~K. Dubey. 2019.
\newblock A study of bfloat16 for deep learning training.
\newblock \emph{ArXiv}, abs/1905.12322.

\bibitem[{Kwon et~al.(2023)Kwon, Li, Zhuang, Sheng, Zheng, Yu, Gonzalez, Zhang, and Stoica}]{10.1145/3600006.3613165}
Woosuk Kwon, Zhuohan Li, Siyuan Zhuang, Ying Sheng, Lianmin Zheng, Cody~Hao Yu, Joseph Gonzalez, Hao Zhang, and Ion Stoica. 2023.
\newblock Efficient memory management for large language model serving with pagedattention.
\newblock In \emph{Proceedings of the 29th Symposium on Operating Systems Principles}, SOSP '23, page 611–626, New York, NY, USA. Association for Computing Machinery.

\bibitem[{Li et~al.(2024{\natexlab{a}})Li, Zhou, Fu, Wang, Roth, and Chen}]{li2024famicom}
Bangzheng Li, Ben Zhou, Xingyu Fu, Fei Wang, Dan Roth, and Muhao Chen. 2024{\natexlab{a}}.
\newblock Famicom: Further demystifying prompts for language models with task-agnostic performance estimation.
\newblock \emph{arXiv preprint arXiv:2406.11243}.

\bibitem[{Li et~al.(2024{\natexlab{b}})Li, Zhou, Wang, Fu, Roth, and Chen}]{li2024deceptive}
Bangzheng Li, Ben Zhou, Fei Wang, Xingyu Fu, Dan Roth, and Muhao Chen. 2024{\natexlab{b}}.
\newblock Deceptive semantic shortcuts on reasoning chains: How far can models go without hallucination?
\newblock In \emph{Proceedings of the 2024 Conference of the North American Chapter of the Association for Computational Linguistics: Human Language Technologies (Volume 1: Long Papers)}, pages 7668--7681.

\bibitem[{Li et~al.(2023)Li, Cheng, Zhao, Nie, and Wen}]{li-etal-2023-halueval}
Junyi Li, Xiaoxue Cheng, Xin Zhao, Jian-Yun Nie, and Ji-Rong Wen. 2023.
\newblock {H}alu{E}val: A large-scale hallucination evaluation benchmark for large language models.
\newblock In \emph{Proceedings of the 2023 Conference on Empirical Methods in Natural Language Processing}, pages 6449--6464, Singapore. Association for Computational Linguistics.

\bibitem[{Lin et~al.(2022)Lin, Hilton, and Evans}]{lin-etal-2022-truthfulqa}
Stephanie Lin, Jacob Hilton, and Owain Evans. 2022.
\newblock {T}ruthful{QA}: Measuring how models mimic human falsehoods.
\newblock In \emph{Proceedings of the 60th Annual Meeting of the Association for Computational Linguistics (Volume 1: Long Papers)}, pages 3214--3252, Dublin, Ireland. Association for Computational Linguistics.

\bibitem[{Lin et~al.(2024)Lin, Gou, Gong, Liu, Shen, Xu, Lin, Yang, Jiao, Duan et~al.}]{lin2024rho}
Zhenghao Lin, Zhibin Gou, Yeyun Gong, Xiao Liu, Yelong Shen, Ruochen Xu, Chen Lin, Yujiu Yang, Jian Jiao, Nan Duan, et~al. 2024.
\newblock Rho-1: Not all tokens are what you need.
\newblock \emph{arXiv preprint arXiv:2404.07965}.

\bibitem[{Liu et~al.(2024)Liu, Lin, Hewitt, Paranjape, Bevilacqua, Petroni, and Liang}]{liu2024lost}
Nelson~F Liu, Kevin Lin, John Hewitt, Ashwin Paranjape, Michele Bevilacqua, Fabio Petroni, and Percy Liang. 2024.
\newblock Lost in the middle: How language models use long contexts.
\newblock \emph{Transactions of the Association for Computational Linguistics}, 12:157--173.

\bibitem[{Loshchilov and Hutter(2019)}]{loshchilov2018decoupled}
Ilya Loshchilov and Frank Hutter. 2019.
\newblock Decoupled weight decay regularization.
\newblock In \emph{International Conference on Learning Representations}.

\bibitem[{Luo et~al.(2023)Luo, Sun, Xu, Zhao, Lou, Tao, Geng, Lin, Chen, and Zhang}]{luo2023wizardmath}
Haipeng Luo, Qingfeng Sun, Can Xu, Pu~Zhao, Jianguang Lou, Chongyang Tao, Xiubo Geng, Qingwei Lin, Shifeng Chen, and Dongmei Zhang. 2023.
\newblock Wizardmath: Empowering mathematical reasoning for large language models via reinforced evol-instruct.
\newblock \emph{arXiv preprint arXiv:2308.09583}.

\bibitem[{Luo et~al.(2024)Luo, Xu, Zhao, Sun, Geng, Hu, Tao, Ma, Lin, and Jiang}]{luo2024wizardcoder}
Ziyang Luo, Can Xu, Pu~Zhao, Qingfeng Sun, Xiubo Geng, Wenxiang Hu, Chongyang Tao, Jing Ma, Qingwei Lin, and Daxin Jiang. 2024.
\newblock Wizardcoder: Empowering code large language models with evol-instruct.
\newblock In \emph{The Twelfth International Conference on Learning Representations}.

\bibitem[{McNemar(1947)}]{McNemar1947}
Quinn McNemar. 1947.
\newblock Note on the sampling error of the difference between correlated proportions or percentages.
\newblock \emph{Psychometrika}, 12(2):153--157.

\bibitem[{Pace et~al.(2024)Pace, Mallinson, Malmi, Krause, and Severyn}]{pace2024west}
Aliz{\'e}e Pace, Jonathan Mallinson, Eric Malmi, Sebastian Krause, and Aliaksei Severyn. 2024.
\newblock West-of-n: Synthetic preference generation for improved reward modeling.
\newblock \emph{arXiv preprint arXiv:2401.12086}.

\bibitem[{Paster et~al.(2024)Paster, Santos, Azerbayev, and Ba}]{paster2024openwebmath}
Keiran Paster, Marco~Dos Santos, Zhangir Azerbayev, and Jimmy Ba. 2024.
\newblock Openwebmath: An open dataset of high-quality mathematical web text.
\newblock In \emph{The Twelfth International Conference on Learning Representations}.

\bibitem[{Radford et~al.(2018)Radford, Narasimhan, Salimans, and Sutskever}]{radford2018improving}
Alec Radford, Karthik Narasimhan, Tim Salimans, and Ilya Sutskever. 2018.
\newblock Improving language understanding with unsupervised learning.

\bibitem[{Rajbhandari et~al.(2020)Rajbhandari, Rasley, Ruwase, and He}]{10.5555/3433701.3433727}
Samyam Rajbhandari, Jeff Rasley, Olatunji Ruwase, and Yuxiong He. 2020.
\newblock Zero: memory optimizations toward training trillion parameter models.
\newblock In \emph{Proceedings of the International Conference for High Performance Computing, Networking, Storage and Analysis}, SC '20. IEEE Press.

\bibitem[{Rasley et~al.(2020)Rasley, Rajbhandari, Ruwase, and He}]{10.1145/3394486.3406703}
Jeff Rasley, Samyam Rajbhandari, Olatunji Ruwase, and Yuxiong He. 2020.
\newblock Deepspeed: System optimizations enable training deep learning models with over 100 billion parameters.
\newblock In \emph{Proceedings of the 26th ACM SIGKDD International Conference on Knowledge Discovery \& Data Mining}, KDD '20, page 3505–3506, New York, NY, USA. Association for Computing Machinery.

\bibitem[{Rescorla(2024)}]{sep-language-thought}
Michael Rescorla. 2024.
\newblock {The Language of Thought Hypothesis}.
\newblock In Edward~N. Zalta and Uri Nodelman, editors, \emph{The {Stanford} Encyclopedia of Philosophy}, {S}ummer 2024 edition. Metaphysics Research Lab, Stanford University.

\bibitem[{Sahu et~al.(2022)Sahu, Rodriguez, Laradji, Atighehchian, Vazquez, and Bahdanau}]{sahu2022data}
Gaurav Sahu, Pau Rodriguez, Issam Laradji, Parmida Atighehchian, David Vazquez, and Dzmitry Bahdanau. 2022.
\newblock Data augmentation for intent classification with off-the-shelf large language models.
\newblock In \emph{Proceedings of the 4th Workshop on NLP for Conversational AI}, pages 47--57.

\bibitem[{Talmor et~al.(2019)Talmor, Herzig, Lourie, and Berant}]{talmor-etal-2019-commonsenseqa}
Alon Talmor, Jonathan Herzig, Nicholas Lourie, and Jonathan Berant. 2019.
\newblock {C}ommonsense{QA}: A question answering challenge targeting commonsense knowledge.
\newblock In \emph{Proceedings of the 2019 Conference of the North {A}merican Chapter of the Association for Computational Linguistics: Human Language Technologies, Volume 1 (Long and Short Papers)}, pages 4149--4158, Minneapolis, Minnesota. Association for Computational Linguistics.

\bibitem[{Touvron et~al.(2023)Touvron, Martin, Stone, Albert, Almahairi, Babaei, Bashlykov, Batra, Bhargava, Bhosale, Bikel, Blecher, Ferrer, Chen, Cucurull, Esiobu, Fernandes, Fu, Fu, Fuller, Gao, Goswami, Goyal, Hartshorn, Hosseini, Hou, Inan, Kardas, Kerkez, Khabsa, Kloumann, Korenev, Koura, Lachaux, Lavril, Lee, Liskovich, Lu, Mao, Martinet, Mihaylov, Mishra, Molybog, Nie, Poulton, Reizenstein, Rungta, Saladi, Schelten, Silva, Smith, Subramanian, Tan, Tang, Taylor, Williams, Kuan, Xu, Yan, Zarov, Zhang, Fan, Kambadur, Narang, Rodriguez, Stojnic, Edunov, and Scialom}]{touvron2023llama2openfoundation}
Hugo Touvron, Louis Martin, Kevin Stone, Peter Albert, Amjad Almahairi, Yasmine Babaei, Nikolay Bashlykov, Soumya Batra, Prajjwal Bhargava, Shruti Bhosale, Dan Bikel, Lukas Blecher, Cristian~Canton Ferrer, Moya Chen, Guillem Cucurull, David Esiobu, Jude Fernandes, Jeremy Fu, Wenyin Fu, Brian Fuller, Cynthia Gao, Vedanuj Goswami, Naman Goyal, Anthony Hartshorn, Saghar Hosseini, Rui Hou, Hakan Inan, Marcin Kardas, Viktor Kerkez, Madian Khabsa, Isabel Kloumann, Artem Korenev, Punit~Singh Koura, Marie-Anne Lachaux, Thibaut Lavril, Jenya Lee, Diana Liskovich, Yinghai Lu, Yuning Mao, Xavier Martinet, Todor Mihaylov, Pushkar Mishra, Igor Molybog, Yixin Nie, Andrew Poulton, Jeremy Reizenstein, Rashi Rungta, Kalyan Saladi, Alan Schelten, Ruan Silva, Eric~Michael Smith, Ranjan Subramanian, Xiaoqing~Ellen Tan, Binh Tang, Ross Taylor, Adina Williams, Jian~Xiang Kuan, Puxin Xu, Zheng Yan, Iliyan Zarov, Yuchen Zhang, Angela Fan, Melanie Kambadur, Sharan Narang, Aurelien Rodriguez, Robert Stojnic, Sergey Edunov, and Thomas
  Scialom. 2023.
\newblock Llama 2: Open foundation and fine-tuned chat models.
\newblock \emph{arXiv preprint arXiv:2307.09288}.

\bibitem[{Wang et~al.(2023{\natexlab{a}})Wang, Wang, Li, Gao, Yin, and Ren}]{wang-etal-2023-scott}
Peifeng Wang, Zhengyang Wang, Zheng Li, Yifan Gao, Bing Yin, and Xiang Ren. 2023{\natexlab{a}}.
\newblock {SCOTT}: Self-consistent chain-of-thought distillation.
\newblock In \emph{Proceedings of the 61st Annual Meeting of the Association for Computational Linguistics (Volume 1: Long Papers)}, pages 5546--5558, Toronto, Canada. Association for Computational Linguistics.

\bibitem[{Wang and Zhou(2024)}]{wang2024chain}
Xuezhi Wang and Denny Zhou. 2024.
\newblock Chain-of-thought reasoning without prompting.
\newblock \emph{arXiv preprint arXiv:2402.10200}.

\bibitem[{Wang et~al.(2023{\natexlab{b}})Wang, Kordi, Mishra, Liu, Smith, Khashabi, and Hajishirzi}]{wang-etal-2023-self-instruct}
Yizhong Wang, Yeganeh Kordi, Swaroop Mishra, Alisa Liu, Noah~A. Smith, Daniel Khashabi, and Hannaneh Hajishirzi. 2023{\natexlab{b}}.
\newblock Self-instruct: Aligning language models with self-generated instructions.
\newblock In \emph{Proceedings of the 61st Annual Meeting of the Association for Computational Linguistics (Volume 1: Long Papers)}, pages 13484--13508, Toronto, Canada. Association for Computational Linguistics.

\bibitem[{Wei et~al.(2022)Wei, Wang, Schuurmans, Bosma, Xia, Chi, Le, Zhou et~al.}]{wei2022chain}
Jason Wei, Xuezhi Wang, Dale Schuurmans, Maarten Bosma, Fei Xia, Ed~Chi, Quoc~V Le, Denny Zhou, et~al. 2022.
\newblock Chain-of-thought prompting elicits reasoning in large language models.
\newblock \emph{Advances in neural information processing systems}, 35:24824--24837.

\bibitem[{Wei et~al.(2023)Wei, Huang, Lu, Zhou, and Le}]{wei2023simple}
Jerry Wei, Da~Huang, Yifeng Lu, Denny Zhou, and Quoc~V Le. 2023.
\newblock Simple synthetic data reduces sycophancy in large language models.
\newblock \emph{arXiv preprint arXiv:2308.03958}.

\bibitem[{Yang et~al.(2024)Yang, Yang, Zhang, Hui, Zheng, Yu, Li, Liu, Huang, Wei et~al.}]{yang2024qwen2}
An~Yang, Baosong Yang, Beichen Zhang, Binyuan Hui, Bo~Zheng, Bowen Yu, Chengyuan Li, Dayiheng Liu, Fei Huang, Haoran Wei, et~al. 2024.
\newblock Qwen2. 5 technical report.
\newblock \emph{arXiv preprint arXiv:2412.15115}.

\bibitem[{Ye et~al.(2024)Ye, Wang, Huang, Chen, Zhang, Moniz, Gao, Geyer, Huang, Chen, Chawla, and Zhang}]{ye2024justiceprejudicequantifyingbiases}
Jiayi Ye, Yanbo Wang, Yue Huang, Dongping Chen, Qihui Zhang, Nuno Moniz, Tian Gao, Werner Geyer, Chao Huang, Pin-Yu Chen, Nitesh~V Chawla, and Xiangliang Zhang. 2024.
\newblock Justice or prejudice? quantifying biases in llm-as-a-judge.
\newblock \emph{arXiv preprint arXiv:2410.02736}.

\bibitem[{Zelikman et~al.(2024)Zelikman, Harik, Shao, Jayasiri, Haber, and Goodman}]{zelikman2024quiet}
Eric Zelikman, Georges Harik, Yijia Shao, Varuna Jayasiri, Nick Haber, and Noah~D Goodman. 2024.
\newblock Quiet-star: Language models can teach themselves to think before speaking.
\newblock \emph{arXiv preprint arXiv:2403.09629}.

\bibitem[{Zhang et~al.(2024)Zhang, Gangal, Lattimer, and Yang}]{zhang-etal-2024-enhancing-hallucination}
Dongxu Zhang, Varun Gangal, Barrett Lattimer, and Yi~Yang. 2024.
\newblock Enhancing hallucination detection through perturbation-based synthetic data generation in system responses.
\newblock In \emph{Findings of the Association for Computational Linguistics ACL 2024}, pages 13321--13332, Bangkok, Thailand and virtual meeting. Association for Computational Linguistics.

\bibitem[{Zhou et~al.(2021)Zhou, Richardson, Ning, Khot, Sabharwal, and Roth}]{zhou2021temporal}
Ben Zhou, Kyle Richardson, Qiang Ning, Tushar Khot, Ashish Sabharwal, and Dan Roth. 2021.
\newblock Temporal reasoning on implicit events from distant supervision.
\newblock In \emph{Proceedings of the 2021 Conference of the North American Chapter of the Association for Computational Linguistics: Human Language Technologies}, pages 1361--1371.

\bibitem[{Zhou et~al.(2024)Zhou, Zhang, Chen, Yu, Wang, Peng, Roth, and Yu}]{Zhou2023Symbolic}
Ben Zhou, Hongming Zhang, Sihao Chen, Dian Yu, Hongwei Wang, Baolin Peng, Dan Roth, and Dong Yu. 2024.
\newblock Conceptual and unbiased reasoning in language models.
\newblock \emph{arXiv preprint arXiv:2404.00205}.

\bibitem[{Zhu et~al.(2023)Zhu, Qi, Zhang, Long, Lin, and Zhou}]{zhu2023pad}
Xuekai Zhu, Biqing Qi, Kaiyan Zhang, Xinwei Long, Zhouhan Lin, and Bowen Zhou. 2023.
\newblock Pad: Program-aided distillation can teach small models reasoning better than chain-of-thought fine-tuning.
\newblock \emph{arXiv preprint arXiv:2305.13888}.

\end{thebibliography}
\clearpage
\appendix
\section{Prompts for Data Generation}
\label{sec:data-generation}
\begin{tcolorbox}[title=\footnotesize Prompt for \textit{Thought Generation},top=1mm,bottom=1mm]
\scriptsize
Task Instruction: Given certain text, you need to predict the next word of it. Moreover, before your output, you could first give short thoughts about how you infer the next word based on the provided context.\\

Here are five examples for the task:\\
Example 0: \{<ex0>\}\\

Example 1: \{<ex1>\}\\

Example 2: \{<ex2>\}\\

Example 3: \{<ex3>\}\\

Example 4: \{<ex4>\}\\

Now please give me your prediction for the thought and next word based on the following context:\\

\{<context>\}\\ 

Thought: \\
Next Word:
\end{tcolorbox}

\begin{tcolorbox}[title=\footnotesize Prompt for \textit{Consistency Check},top=1mm,bottom=1mm]
\scriptsize
Task Instruction: Given the following certain text, thought for its next word and the gold next word, you need to judge whether the thought for generating the next word is consistent based on the reasoning process and the given text. For consistency, we mean that the thought only needs to generally entail the gold next word in reasoning and does NOT need to be specific on the gold next words.\\

Context: \{<context>\}\\
Thought: \{<thought>\}\\
Gold Next Word: \{<next\_word>\}\\

Now please give me your reasoning and judgement, i.e. True or False, for the consistency of thought and gold next word based on the above information.\\

Reasoning: Let's think step by step. \\
Judgement: 
\end{tcolorbox}

\begin{tcolorbox}[title=\footnotesize Prompt for \textit{Summarization Prompt} (\textit{exact match words}),top=1mm,bottom=1mm]
\scriptsize
Task Instruction: Please modify the following thought into a shorter one within 15 words without changing much of the meaning. The thought is used to help predict the next word of the following context.\\
Context: \{<context>\}\\
Thought: \{<thought>\}\\
Shorter Thought:
\end{tcolorbox}

\begin{tcolorbox}[title=\footnotesize Prompt for \textit{Summarization and Denoising Prompt} (\textit{soft consistent words}),top=1mm,bottom=1mm]
\scriptsize
Task Instruction: Please modify the following thought into a shorter one within 15 words without changing much of the meaning. The thought is used to help predict the next word of the following context. Besides, the gold next word is also given. You should try to shorten the thought based on it.\\ 
Context: \{<context>\}\\
Thought: \{<thought>\}\\
Gold Next Word: \{<gold\_next>\}\\
Shorter Thought:
\end{tcolorbox}

\section{Prompts for Evaluation}
\label{sec:eval-prompts}
\begin{tcolorbox}[title=\footnotesize Prompt for \textit{GSM8K},top=1mm,bottom=1mm]
\scriptsize
Answer the following question.\\

Question: \{<question>\}\\
Answer: Let's think step by step.
\end{tcolorbox}

\begin{tcolorbox}[title=\footnotesize Prompt for \textit{Other MCQ datasets},top=1mm,bottom=1mm]
\scriptsize
Question: Mass-society theory suggests that:\\
Options: A. the content of the media is determined by market forces\\
B. the subordinate classes are dominated by the ideology of the ruling class\\
C. the media manipulate 'the masses' as vulnerable, passive consumers\\
D. audiences make selective interpretations of media messages\\

You should ONLY choose the letters from the options as your final answer.\\
Response: Let's think step by step. Mass-society theory suggests that media content is used to manipulate the masses as passive consumers, who are vulnerable to external influence. Option C reflects this idea, as it aligns with the theory's view that media has the power to control and shape the behavior of large, undifferentiated audiences. The theory sees individuals as passive, easily influenced, and lacking in critical engagement with media content, thus being susceptible to manipulation. So the answer is C.\\

Question: What was GDP per capita in the United States in 1850 when adjusting for inflation and PPP in 2011 prices?\\
Options: A. About \$300\\
B. About \$3k\\
C. About \$8k\\
D. About \$15k

You should ONLY choose the letters from the options as your final answer.\\
Response: Let's think step by step. To estimate GDP per capita in 1850 using inflation-adjusted and PPP-adjusted 2011 prices, historical economic data suggests that early industrial societies like the United States had modest per capita income compared to modern standards. GDP per capita around this period was likely in the range of a few thousand dollars when adjusted to 2011 prices. Option B,``About \$3k'' aligns with historical estimates of the U.S. economy in the mid-19th century, reflecting moderate economic development during this era. So the answer is B.\\

Question: Which common public relations tactic involves sending journalists on visits to appropriate locations?\\
Options: A. Media release\\
B. Media tour\\
C. Press room\\
D.Promotional days/weeks\\

You should ONLY choose the letters from the options as your final answer.\\
Response: Let's think step by step. A media tour involves sending journalists to relevant locations to give them firsthand experience of a product, service, or event. This tactic helps create more informed and engaging reports by providing journalists with direct exposure to the subject. Option B is correct because a media tour specifically entails organizing trips or visits for journalists to gain a deeper understanding and coverage of a particular topic. Other options, like media releases, do not involve physical visits. So the answer is B.\\

Question: \{<question>\}\\
Options: \{<choices>\} \\

You should ONLY choose the letter from the options as your final answer.\\
Response: Let's think step by step.
\end{tcolorbox}
The above 3-shot examples are randomly chosen from MMLU \cite{hendryckstest2021} test set. For binary classification benchmarks, we transform them into MCQ dataset.

\section{Training and Inference Details}
\label{sec:training}
For training, we use the AdamW optimizer with a learning rate of 2$e-$5 and weight decay of 0. We use 3\% as the warmup ratio and a linear learning rate scheduler. We use a maximum sequence length of 3072 for \method{}-NoDeN and 2048 for \method{} during training.
We use meta-tokens to wrap the thoughts of words, and initialize the embeddings of the meta-tokens with the embeddings corresponding to the em dash ``\verb!---!'', which often appears in text data to denote a pause or
thought. Specifically, we use \texttt{<\method{}>} and \texttt{</\method{}>} to wrap thoughts of words. 
To enable efficient finetuning of LLMs, we use the DeepSpeed library \cite{10.1145/3394486.3406703} and ZeRO stage 2 optimizer \cite{10.5555/3433701.3433727}. All models are trained with BFloat16 \cite{Kalamkar2019ASO} mixed precision for stability.  During inference, for models trained on \method{}-NoDeN, we use a maximum token length of 2048, given that the thoughts are generally longer in \method{}-NoDeN. For models trained on \method{}, we use the maximum token length of 512, given that the thoughts are short. 
All experiments are conducted on 8 NVIDIA A100 GPUs.

\section{Exemplars and Further Case Study for \method{}}
\label{sec:example}
\paragraph{Mathematical Reasoning (GSM8K):} As demonstrated in Fig.~\ref{fig:GSM_APDX}, examples from \textit{GSM8K} focus on multi-step mathematical reasoning. In each case, the \method{} approach arrives at the correct conclusion, while the \textsc{Raw} models suffer from unit conversion errors or misinterpretations of the problem. For instance, in Example 1, the \method{} model correctly converts Topher's shoe length from feet and inches to inches (8 feet 4 inches equals 100 inches) and sets up the appropriate equation to solve for Bobby's shoe length in the ToW generation process. Conversely, the \textsc{Raw} model incorrectly converts the length to 104 inches and arrives at an incorrect answer.

\paragraph{Science Reasoning (ARC-Challenge):} Fig.~\ref{fig:ARC_APDX} presents an additional example from the \textit{ARC-Challenge} dataset, which includes multiple-choice science questions. In this example, the \method{} underscores the context of the given question, which enables the model to arrive at the correct answer. Without highlighting the question context, the \textsc{Raw} model loses track of what question to answer. As a result, although its prediction is correct in the sense that new energy-efficient lightbulbs can save electricity bills, it does not answer the question regarding the environmental impact.

\paragraph{Commonsense Reasoning (CSQA):} Examples from the \textit{CSQA} dataset, illustrated in Fig.~\ref{fig:CSQA_APDX}, highlight a common issue in \textsc{Raw} outputs, where the decoding process generates irrelevant and unfocused information without reaching a conclusion. The \method{} counterparts draw logical connections that help avoid this pitfall. For example, when asked where traveling clothes are often kept, the \method{} model logically infers that they are kept in a suitcase, while the \textsc{Raw} model provides repetitive and non-conclusive information. Similarly, in determining where one might find a snake in tall grass, the \method{} model correctly identifies a field, whereas the \textsc{Raw} model fails to provide a coherent answer.

\paragraph{Truthful Reasoning (TruthfulQA):} Fig.~\ref{fig:TruthfulQA_APDX} shows an additional example from the \textit{TruthfulQA} dataset. In this example, regarding the location of Walt Disney's body, the \method{} model accurately states that Walt Disney was cremated after his death, while the \textsc{Raw} model makes false claims and arrives at the incorrect conclusion.

\clearpage

\begin{tcolorbox}[width=\textwidth, colframe=black, colback=white, boxrule=1pt]
\textbf{\large{Example 1:}} \\
\begin{tcolorbox}[width=\textwidth, colback=yellow!10, boxrule=1pt]
\small{\textbf{Question:} Topher, the green giant, wears enormous shoes. The length of one of his shoes is 10 inches longer than 9 times the length of one of Bobby’s shoes. If the length of one of Topher’s shoes is 8-feet and 4-inches, how long, in inches, is one of Bobby’s shoes? \\ \textbf{{Answer: 10}}}
\end{tcolorbox}

\begin{minipage}[t]{0.48\textwidth}
\begin{tcolorbox}[colback=green!10, valign=top, equal height group=ex1]
\textbf{\method{}} \cmark: \\
\footnotesize
\textcolor{red}{\texttt{<\method{}>}} The context involves solving a math problem about shoe lengths, starting with Topher's. \textcolor{red}{\texttt{</\method{}>}} Topher's shoe is \textcolor{red}{\texttt{<\method{}>}} 8-feet and 4-inches, which is 100 inches. \textcolor{red}{\texttt{</\method{}>}} According to the problem, the length of one of Topher’s shoes is 10 inches longer than 9 times the length of one of Bobby’s shoes. So, we can write an equation: 100 = 9x + 10. Now, we need to solve for x. \textcolor{red}{\texttt{<\method{}>}} 90 = 9x. Next, we will divide each side by 9. \textcolor{red}{\texttt{</\method{}>}} 10 = x. So, Bobby's shoe is \textbf{10} inches long.
\end{tcolorbox}
\end{minipage}%
\hfill
\begin{minipage}[t]{0.48\textwidth}
\begin{tcolorbox}[colback=red!10, valign=top, equal height group=ex1]
\textbf{\textsc{Raw}} \xmark: \\
\footnotesize
Topher's shoe is 8-feet and 4-inches. That is 8 times 12 inches plus 4 inches, which is 104 inches. The question tells us that the length of Topher's shoe is 9 times the length of Bobby's shoe plus 10 inches. So, we can write an equation. 104 = 9B + 10. We need to solve for B. 94 = 9B. 94/9 = B. 10.4444... = B. The length of Bobby's shoe is \textbf{10.4444}... inches.
\end{tcolorbox}
\end{minipage}

\vspace{15pt}

\textbf{\large{Example 2:}} \\
\begin{tcolorbox}[width=\textwidth, colback=yellow!10, boxrule=1pt]
\small{\textbf{Question:} Bethany can run 10 laps on the track in one hour. Trey can run 4 more laps than Bethany. Shaelyn can run half as many laps as Trey. Quinn can run 2 fewer laps than Shaelyn. How many more laps can Bethany run compared to Quinn?} \\
\textbf{Answer: 5}
\end{tcolorbox}

\begin{minipage}[t]{0.48\textwidth}
\begin{tcolorbox}[colback=green!10, valign=top, equal height group=ex2]
\textbf{\method{}} \cmark: \\
\footnotesize
\textcolor{red}{\texttt{<\method{}>}} Bethany runs 10 laps. Trey runs 4 more laps than Bethany, so he runs 14 laps. Shaelyn runs half as many laps as Trey, so she runs 7 laps. Quinn runs 2 fewer laps than Shaelyn, so he runs 5 laps. \textcolor{red}{\texttt{</\method{}>}} Bethany runs \textbf{5} more laps than Quinn.
\end{tcolorbox}
\end{minipage}%
\hfill
\begin{minipage}[t]{0.48\textwidth}
\begin{tcolorbox}[colback=red!10, valign=top, equal height group=ex2]
\textbf{\textsc{Raw}} \xmark: \\
\footnotesize
Bethany can run 10 laps in one hour. That means she can run 1 lap in 6 minutes. Trey can run 1 lap in 4.5 minutes. Shaelyn can run 1 lap in 2.25 minutes. Quinn can run 1 lap in 1.5 minutes. Bethany can run \textbf{1} more lap than Quinn.
\end{tcolorbox}
\end{minipage}
\end{tcolorbox}
\begin{tcolorbox}[width=\textwidth, colframe=white, colback=white, boxrule=0pt]
\centering
\captionof{figure}{The comparison of \method{} vs. \textsc{Raw} outputs on multiple examples from the \textit{GSM8K} dataset. }
\label{fig:GSM_APDX}
\end{tcolorbox}
\clearpage

\newpage
\begin{tcolorbox}[width=\textwidth, colframe=black, colback=white, boxrule=1pt]

\textbf{\large{Example 1:}} \
\begin{tcolorbox}[width=\textwidth, colback=yellow!10, boxrule=1pt]
\small{\textbf{Question:} A school replaced all of their light bulbs with a new type of light bulb. The new light bulbs use less energy. How will the new light bulbs most likely help the environment? Options: A. The amount of light pollution will decrease. B. The air conditioner at the school will run less often. C. Fewer natural resources will be used to make electricity. D. Money will be saved on the electricity bill for the school.} \\
\textbf{Answer: C}
\end{tcolorbox}
\begin{minipage}[t]{0.48\textwidth}
\begin{tcolorbox}[colback=green!10, valign=top, equal height group=ex11]
\textbf{\method{}} \cmark: \\
\footnotesize
Replacing \textcolor{red}{\texttt{<\method{}>}} The context discusses replacing light bulbs to conserve energy and reduce environmental impact. \textcolor{red}{\texttt{</\method{}>}} light \textcolor{red}{\texttt{<\method{}>}} The context discusses replacing light bulbs to conserve energy and reduce environmental impact. \textcolor{red}{\texttt{</\method{}>}} bulbs with more efficient ones will save energy and reduce pollution. Option C is correct because less energy means less pollution from power plants. So the answer is \textbf{C}.
\end{tcolorbox}
\end{minipage}%
\hfill
\begin{minipage}[t]{0.48\textwidth}
\begin{tcolorbox}[colback=red!10, valign=top, equal height group=ex11]
\textbf{\textsc{Raw}} \xmark: \\
\footnotesize
The new light bulbs will use less energy, so the school will save money on its electricity bill. This will also mean that fewer natural resources will be used to make electricity, which is good for the environment. Option D is correct because it is the most specific answer. So the answer is \textbf{D}.
\end{tcolorbox}
\end{minipage}

\vspace{15pt}

\end{tcolorbox}
\begin{tcolorbox}[width=\textwidth, colframe=white, colback=white, boxrule=0pt]
\centering
\captionof{figure}{The comparison of \method{} vs. \textsc{Raw} outputs on an additional example from the \textit{ARC-Challenge} dataset.}
\label{fig:ARC_APDX}
\end{tcolorbox}
\clearpage 

\newpage
\begin{tcolorbox}[width=\textwidth, colframe=black, colback=white, boxrule=1pt]
\vspace{15pt}

\textbf{\large{Example 1:}} \\
\begin{tcolorbox}[width=\textwidth, colback=yellow!10, boxrule=1pt]
\small{\textbf{Question:} Where are traveling clothes often kept? Options: A. suitcase B. bedroom C. closet D. drawer E. dresser} \\
\textbf{Answer: A}
\end{tcolorbox}

\begin{minipage}[t]{0.48\textwidth}
\begin{tcolorbox}[colback=green!10, valign=top, equal height group=ex5]
\textbf{\method{}} \cmark: \\
\footnotesize
Clothes \textcolor{red}{\texttt{<\method{}>}} The context asks about clothing storage, suggesting "kept" as the next logical word. \textcolor{red}{\texttt{</\method{}>}} kept in a \textcolor{red}{\texttt{<\method{}>}} The context asks where traveling clothes are stored, suggesting "suitcase" as the answer. \textcolor{red}{\texttt{</\method{}>}} suitcase are \textcolor{red}{\texttt{<\method{}>}} The context asks where traveling clothes are stored, suggesting "suitcase" as the answer. \textcolor{red}{\texttt{</\method{}>}} traveling \textcolor{red}{\texttt{<\method{}>}} The context asks where traveling clothes are stored, suggesting "clothes" as the next word. \textcolor{red}{\texttt{</\method{}>}} clothes. So the answer is \textbf{A}.
\end{tcolorbox}
\end{minipage}%
\hfill
\begin{minipage}[t]{0.48\textwidth}
\begin{tcolorbox}[colback=red!10, valign=top, equal height group=ex5]
\textbf{\textsc{Raw}} \xmark: \\
\footnotesize
1. A suitcase is a container used to carry clothes when traveling. It is usually made of hard material like leather or plastic and has a handle for easy carrying. 2. A bedroom is a room in a house or apartment where people sleep and rest. It is usually furnished with a bed, a closet, and other pieces of furniture. 3. A closet is a small room or a part of a room where clothes are stored... \textcolor{blue}{(continues listing)}
\end{tcolorbox}
\end{minipage}

\vspace{15pt}

\textbf{\large{Example 2:}} \\
\begin{tcolorbox}[width=\textwidth, colback=yellow!10, boxrule=1pt]
\small{\textbf{Question:} Where can you find a snake in tall grass? Options: A. tree B. in a jar C. pet shops D. field E. tropical forest} \\
\textbf{Answer: D}
\end{tcolorbox}

\begin{minipage}[t]{0.48\textwidth}
\begin{tcolorbox}[colback=green!10, valign=top, equal height group=ex6]
\textbf{\method{}} \cmark: \\
\footnotesize
Snakes live in \textcolor{red}{\texttt{<\method{}>}} The context asks where snakes live, suggesting "tall grass" as a likely answer. \textcolor{red}{\texttt{</\method{}>}} tall \textcolor{red}{\texttt{<\method{}>}} The context asks where snakes live, suggesting "grass" as the next word. \textcolor{red}{\texttt{</\method{}>}} grass. So the answer is \textbf{D}.
\end{tcolorbox}
\end{minipage}%
\hfill
\begin{minipage}[t]{0.48\textwidth}
\begin{tcolorbox}[colback=red!10, valign=top, equal height group=ex6]
\textbf{\textsc{Raw}} \xmark: \\
\footnotesize
1. A snake is a reptile that lives in the grass. 2. A snake is a reptile that lives in the grass. 3. A snake is a reptile that lives in the grass... \textcolor{blue}{(repeats many times)}
\end{tcolorbox}
\end{minipage}

\end{tcolorbox}

\begin{tcolorbox}[width=\textwidth, colframe=white, colback=white, boxrule=0pt]
\centering
\captionof{figure}{The comparison of \method{} vs. \textsc{Raw} outputs on multiple examples from the \textit{CSQA} dataset.}
\label{fig:CSQA_APDX}
\end{tcolorbox}
\clearpage 

\newpage

\begin{tcolorbox}[width=\textwidth, colframe=black, colback=white, boxrule=1pt]
\vspace{15pt}




\textbf{\large{Example 1:}} \\
\begin{tcolorbox}[width=\textwidth, colback=yellow!10, boxrule=1pt]
\small{\textbf{Question:} Where is Walt Disney's body? Options: A. Walt Disney's body was cremated after his death. B. Walt Disney's body is in Hollywood. C. Walt Disney's body is in Anaheim, California. D. Walt Disney's body is in suspended animation. E. Walt Disney's body is in Disneyland. F. Walt Disney's body is beneath the Pirates of the Caribbean ride. G. Walt Disney's body is in a secret bunker. H. Walt Disney's body is cryonically frozen. I. There are claims that Walt Disney's body was interred in Glendale, California, but he was actually cryonically frozen. J. Walt Disney's body is in Florida. K. Walt Disney's body is cryogenically frozen.} \\
\textbf{Answer: A}
\end{tcolorbox}

\begin{minipage}[t]{0.48\textwidth}
\begin{tcolorbox}[colback=green!10, valign=top, equal height group=ex4]
\textbf{\method{}} \cmark: \\
\footnotesize
Walt \textcolor{red}{\texttt{<\method{}>}} The context lists locations of Walt Disney's body, suggesting the next word is "Disney." \textcolor{red}{\texttt{</\method{}>}} Disney's body was cremated after his death. The other options are false. So the answer is \textbf{A}.
\end{tcolorbox}
\end{minipage}%
\hfill
\begin{minipage}[t]{0.48\textwidth}
\begin{tcolorbox}[colback=red!10, valign=top, equal height group=ex4]
\textbf{\textsc{Raw}} \xmark: \\
\footnotesize
1. A. Walt Disney's body was cremated after his death. FALSE. 2. B. Walt Disney's body is in Hollywood. FALSE. 3. C. Walt Disney's body is in Anaheim, California. FALSE. 4. D. Walt Disney's body is in suspended animation. FALSE. 5. E. Walt Disney's body is in Disneyland. FALSE. 6. F. Walt Disney's body is beneath the Pirates of the Caribbean ride. FALSE. 7. G. Walt Disney's body is in a secret bunker. FALSE. 8. H. Walt Disney's body is cryonically frozen. FALSE. 9. I. There are claims that Walt Disney's body was interred in Glendale, California, but he was actually cryonically frozen. FALSE. 10. J. Walt Disney's body is in Florida. FALSE. 11. K. Walt Disney's body is cryogenically frozen. FALSE. So the answer is \textbf{K}.
\end{tcolorbox}
\end{minipage}

\end{tcolorbox}
\begin{tcolorbox}[width=\textwidth, colframe=white, colback=white, boxrule=0pt]
\centering
\captionof{figure}{The comparison of \method{} vs. \textsc{Raw} output on an additional example from the \textit{TruthfulQA} dataset. 
}
\label{fig:TruthfulQA_APDX}
\end{tcolorbox}
\clearpage

\end{document}